\documentclass[runningheads]{llncs}

\usepackage{eccv}

\usepackage{eccvabbrv}

\usepackage{graphicx}
\usepackage{booktabs}

\usepackage{float}
\usepackage{caption}
\usepackage{wrapfig}
\usepackage{xspace}

\usepackage{booktabs}
\usepackage{multirow}
\usepackage{array}
\usepackage{makecell} % Fixed the missing package from image_bcb63d.png
\usepackage{rotating}
\usepackage[table]{xcolor} % 표 하이라이트를 위한 패키지
\usepackage{graphicx, makecell, booktabs}

\usepackage{marvosym}

\usepackage[accsupp]{axessibility}  % Improves PDF readability for those with disabilities.

\usepackage{hyperref}

\usepackage{orcidlink}

\begin{document}

% ---------------------------------------------------------------
% TODO REVIEW: Replace with your title
\title{
\textbf{FastSTAR: 
Spatiotemporal Token Pruning for Efficient Autoregressive Video Synthesis}
}

% TODO REVIEW: If the paper title is too long for the running head, you can set
% an abbreviated paper title here. If not, comment out.
\titlerunning{FastSTAR}

% TODO FINAL: Replace with your author list. 
% Include the authors' OCRID for the camera-ready version, if at all possible.
\author{Sungwoong Yune\thanks{Equal contribution}\orcidlink{0009-0007-5440-9265} \and
Suheon Jeong\protect\footnotemark[1] \and
Joo-Young Kim\textsuperscript{(\Letter)}}

% TODO FINAL: Replace with an abbreviated list of authors.
\authorrunning{S.Yune et al.}
% First names are abbreviated in the running head.
% If there are more than two authors, 'et al.' is used.

% TODO FINAL: Replace with your institution list.
\institute{Korea Advanced Institute of Science and Technology \\
\email{\{imwooong, sh.jeong, jooyoung1203\}@kaist.ac.kr}}

\maketitle

\newcommand{\methodname}{FastSTAR\xspace}

\vspace{-0.2in}
\begin{figure}[H]
  \centering
  \includegraphics[width=0.9\textwidth]{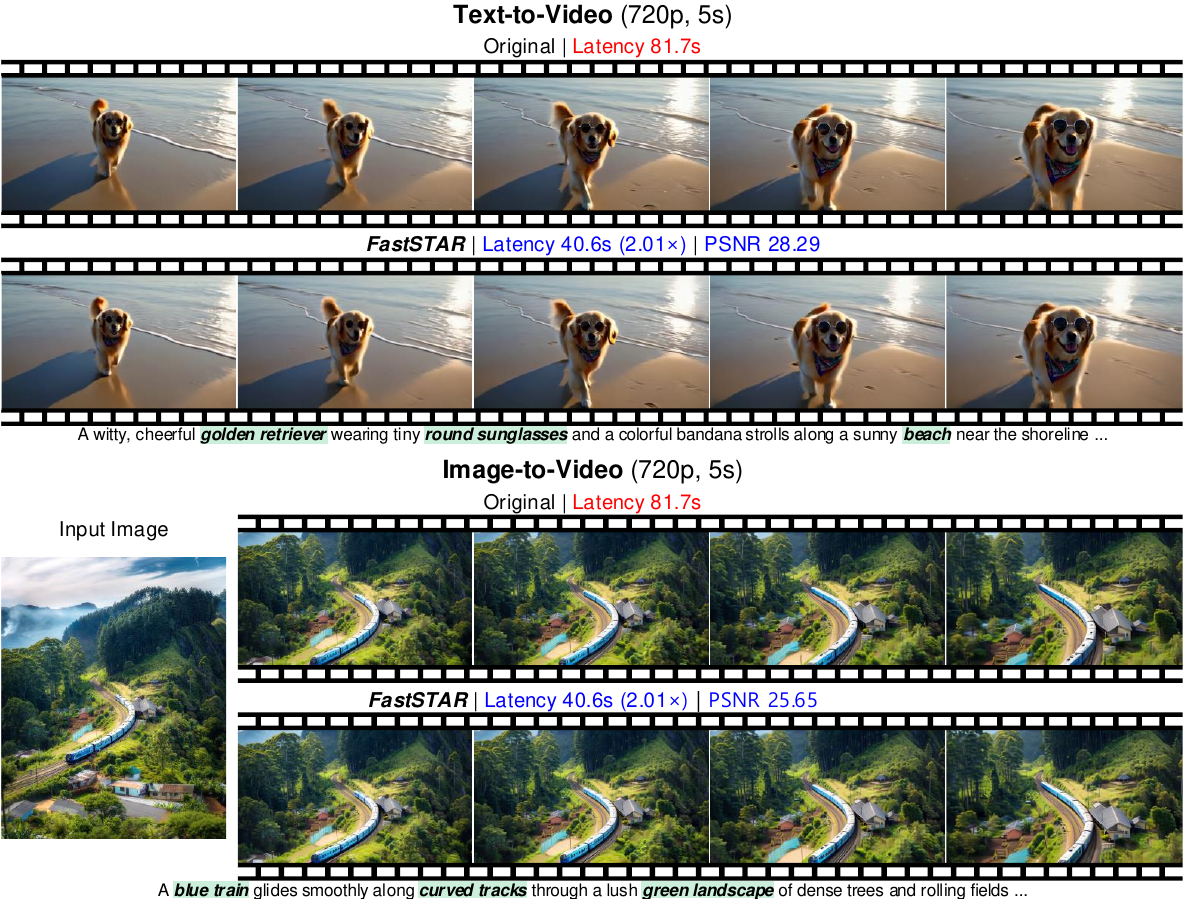}
  \caption{\methodname achieves a 2.01$\times$ end-to-end speedup on a single H100 GPU for T2V and I2V tasks, maintaining high fidelity with PSNR of 28.29 and 25.65, respectively.}
  \vspace{-0.2in}
  \label{abs_1}
\end{figure}

\begin{abstract}
Visual Autoregressive modeling (VAR) has emerged as a highly efficient alternative to diffusion-based frameworks, achieving comparable synthesis quality. However, as this paradigm extends to Spacetime Autoregressive modeling (STAR) for video generation, scaling resolution and frame counts leads to a "token explosion" that creates a massive computational bottleneck in the final refinement stages. To address this, we propose \textbf{\methodname}, a training-free acceleration framework designed for high-quality video generation. Our core method, \textbf{Spatiotemporal Token Pruning}, identifies essential tokens by integrating two specialized terms: (1) \textit{Spatial similarity}, which evaluates structural convergence across hierarchical scales to skip computations in regions where further refinement becomes redundant, and (2) \textit{Temporal similarity}, which identifies active motion trajectories by assessing feature-level variations relative to the preceding clip. Combined with a \textbf{Partial Update} mechanism, \methodname ensures that only non-converged regions are refined, maintaining fluid motion while bypassing redundant computations. Experimental results on InfinityStar demonstrate that \methodname achieves up to a $2.01\times$ speedup with a PSNR of 28.29 and less than 1\% performance degradation, proving a superior efficiency-quality trade-off for STAR-based video synthesis.

\end{abstract}
\section{Introduction}

\begin{figure}[!htbp]
    \centering
    \vspace{-0.3in}
    \includegraphics[width=\textwidth]{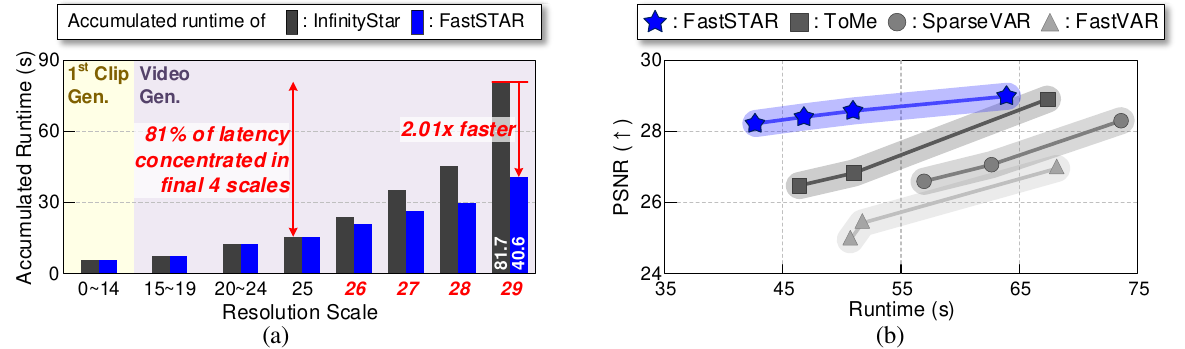}
    \vspace{-0.2in}
    \caption{
    (a) Latency breakdown for generating a 720p video (5s, 81 frames), comparing InfinityStar with \methodname. (b) Latency vs.\ PSNR trade-off curves for T2V synthesis. \methodname establishes the Pareto frontier, consistently surpassing existing baselines.
    }
    \label{intro_1_horizontal}
    \vspace{-0.2in}
    % \vspace{-0.06in}
\end{figure}

Visual Autoregressive modeling (VAR)~\cite{var, infinity} has recently emerged as a powerful paradigm for visual generation, offering superior inference efficiency by redefining image synthesis as a coarse-to-fine next-scale prediction task. Unlike diffusion models~\cite{pixart_alpha, pixart_sigma, pixart_delta, stable_diffusion} that synthesize content through the iterative denoising of random noise~\cite{ddim, ddpm, dpmsolver}, VAR-based frameworks progressively accumulate multi-scale residual feature maps. Building on this success, the recent Spacetime Autoregressive modeling (STAR)~\cite{infinitystar} utilizes a 3D-VAE~\cite{3dvae} to tokenize video data into a spacetime pyramid, maintaining robust temporal consistency. However, the addition of the temporal dimension $T$ escalates the computational complexity of the attention layer from $\mathcal{O}(H^{2}W^{2})$ to $\mathcal{O}(T^{2}H^{2}W^{2})$, where $H$ and $W$ denote the spatial resolution of the token map. Our profiling reveals that this quadratic growth manifests as a critical computational imbalance, where the last 4 out of the total resolution scales account for a disproportionate 81\% of the total inference latency (Fig.~\ref{intro_1_horizontal}(a)).

To address this "token explosion", existing reduction methods~\cite{sparsevar, fastvar, tome} suffer from distinct architectural limitations when applied to STAR. First, image-centric metric evaluations lead to the inaccurate identification of essential tokens, as they struggle to effectively detect the intricate temporal dynamics and motion trajectories inherent in video. Second, structural mismatches in token merging distort discrete feature distributions, triggering an error feedback loop that propagates across subsequent scales and spreads spatially as resolution increases. To overcome these challenges, scale-wise spectral analysis (\S~\ref{method:fourier}) demonstrates that while low-frequency global structures stabilize in the early stages, high-frequency details necessitate continuous updates until the final scale.

\begin{figure}[t]
    \centering
    \includegraphics[width=\textwidth]{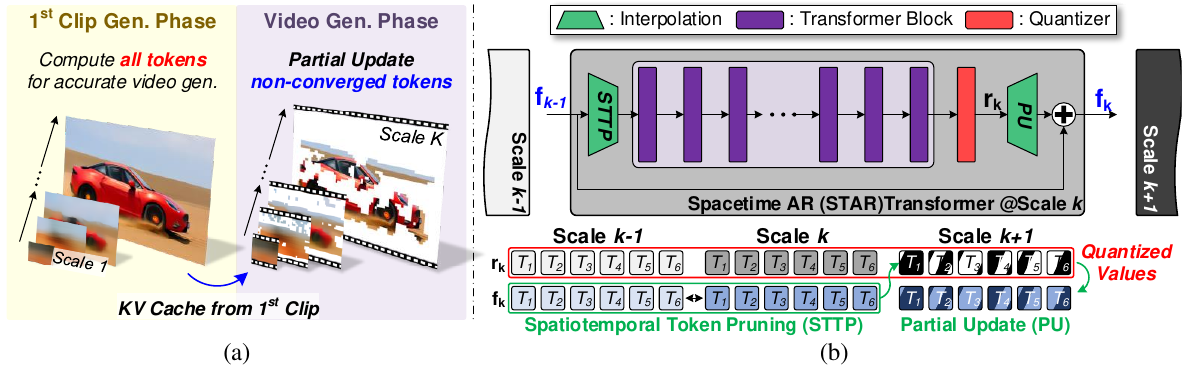}
    \vspace{-0.2in}
    \caption{
    (a) Overview of the \methodname framework. (b) Overall Mechanism of \methodname; Spatiotemporal Token Pruning identifies converged regions to completely skip transformer blocks, while Partial Update preserves structural integrity.
    }
    \label{intro_2}
    \vspace{-0.2in}
    % \vspace{-0.06in}
\end{figure}

To leverage these insights, we propose \textbf{\methodname}, a training-free acceleration framework specifically tailored for high-quality video synthesis (Fig.~\ref{intro_2}). Unlike conventional token merging~\cite{tome, vidtome} or cached-token restoration~\cite{scaleKV} that often distort latent representations and degrade visual fidelity in the VAR paradigm, \methodname adopts a pruning-over-merging strategy to preserve structural integrity. Our framework consists of two core components:
\begin{itemize}
    \item \textbf{Spatiotemporal Token Pruning (STTP):} It identifies non-converged tokens by integrating spatial and temporal similarity. The spatial term evaluates structural convergence across scales, while the temporal term captures motion dynamics via the causal consistency of STAR-based generation.
    \item \textbf{Partial Update (PU):} This mechanism prevents updates in pruned regions, confining transformer refinement exclusively to non-converged tokens. Restricting refinement to these active areas eliminates redundant processing while safeguarding the structural integrity of the cumulative feature map.
\end{itemize}

\noindent Experimental results on InfinityStar demonstrate that \methodname achieves an end-to-end speedup up to \textbf{2.01$\times$} on a single NVIDIA H100 GPU while maintaining a high PSNR of 28.29. With less than 1\% performance degradation, \methodname establishes a new Pareto frontier in Text-to-Video generation, consistently outperforming existing acceleration methods as shown in Fig.~\ref{intro_1_horizontal}(b). By delivering robust gains across Image-to-Video and Video-to-Video tasks without any additional fine-tuning, our approach provides a practical, scalable solution for state-of-the-art autoregressive video synthesis.
\section{Related Work}

% \subsection{Autoregressive Video Generation}
\noindent\textbf{Autoregressive Video Generation.}
Early autoregressive (AR)-based video generators treat synthesis as sequential next-token prediction over discrete visual representations~\cite{emu3, videogpt, magvit}, which inherently discards spatial structure by serializing 3D content into a 1D sequence. To address this, some approaches introduce spatiotemporal-aware positional embeddings~\cite{lumos} or remove vector quantization~\cite{vq1, vq2} entirely with spatial set-by-set prediction~\cite{nova}. Recent research also explores spatiotemporal cubes as a more tightly coupled prediction unit~\cite{nxtframe}. In parallel, VAR-based models reformulate video synthesis as a coarse-to-fine next-scale prediction~\cite{videoar,infinitystar}, achieving synthesis quality comparable to diffusion-based models~\cite{hunyuanvideo, cogvideox} at a fraction of the inference cost.

% \vspace{0.1in}
\medskip

% \subsection{Efficient Image Synthesis}
\noindent\textbf{Efficient Image Synthesis.}
A diverse landscape of efficiency-oriented strategies has been established for both VAR-based and diffusion-based frameworks. Within the VAR-based framework, existing approaches reduce computational cost through (i) scale-wise token selection with cached-token restoration (FastVAR~\cite{fastvar}), (ii) frequency-aware schemes that skip generation steps~\cite{skipvar} and stage-aware schemes that prune or approximate computation in later stages~\cite{stagevar}, and (iii) low-frequency token exclusion in high-resolution stages while preserving fidelity via anchor tokens (SparseVAR~\cite{sparsevar}). Other directions improve efficiency without explicitly dropping tokens, such as compressing KV cache in a scale-aware manner~\cite{scaleKV}, or modifying decoding to share computation across candidates~\cite{code}. In parallel, diffusion-based image synthesis has developed efficiency techniques: step-reduction via fast solvers~\cite{dpmsolver}, as well as runtime optimizations that reduce denoising compute through feature caching~\cite{toca}, similarity-based token pruning~\cite{sito,dato}, and diffusion process-aware token merging~\cite{tomesd,sdtm,toma}.

% \vspace{0.1in}
\medskip

% \subsection{Efficient Video Synthesis}
\noindent\textbf{Efficient Video Synthesis.}
The transition from image to video synthesis significantly amplifies computational demands, necessitating specialized acceleration efforts across both autoregressive and diffusion paradigms. On the AR side, training-free methods exploit spatiotemporal redundancy during sequential decoding~\cite{diagd} or mitigate the growing KV-cache bottleneck via compact cache management~\cite{packcache}. For diffusion-based video synthesis, efficiency has been extensively studied through step-reduction via distillation or consistency training~\cite{videolcm,mcm,animatediff_lightning}, as well as optimizations including feature caching across denoising steps~\cite{fastercache,easycache}, token-level reduction via asymmetric attention pruning~\cite{asymrnr} and cross-frame token merging~\cite{vidtome}, and pruning of redundant temporal attention~\cite{f3pruning}. Despite their effectiveness, existing AR frameworks focus on flat sequential decoding, lacking mechanisms to exploit scale-wise redundancy inherit in STAR-based hierarchical models. Meanwhile, diffusion-centric methods rely on cross-step redundancy within denoising trajectories, a property absent in discrete-index generation, making them not directly transferable.

% These training-free approaches commonly exploit cross-step redundancy in the denoising trajectory, where intermediate features exhibit high similarity across adjacent steps. Despite their effectiveness, such properties are absent in VAR/STAR-style hierarchical refinement over discrete quantized indices, making these methods not directly transferable.

\section{Method}

\subsection{Preliminary}

\noindent \textbf{Visual Autoregressive Modeling.} 
Visual Autoregressive modeling (VAR) reformulates image synthesis as a next-scale prediction task. Within this framework, generation is performed through a hierarchy of discrete token maps $r_1, r_2,$ $\dots, r_{K}$ with progressively increasing resolutions $(h_k, w_k)$. Each token map $r_k$ at a given scale follows a joint probability distribution conditioned on the results of all preceding scales:
\begin{equation}
p(r_1, \dots, r_{K}) = \prod_{k=1}^{K} p(r_k \mid r_1, \dots, r_{k-1})
\end{equation}
This hierarchical structure enables the model to establish global context at lower scales and refine intricate details at higher resolutions.

\medskip

\noindent \textbf{Spatiotemporal Pyramid Modeling.} 
Following recent spacetime autoregressive frameworks, a video is decomposed into $N$ sequential clips $\{c_1, c_2, \dots, c_{N}\}$. The initial clip $c_1$ (at $T=1$) encodes static appearance cues and establishes the spatiotemporal foundation for the sequence. Subsequent clips where $T > 1$ are modeled as 3D volume pyramid with dimensions $(T, h_k, w_k)$ and are generated:
\begin{equation}
p(c_1, \dots, c_{N}) = \prod_{c=1}^{N} \prod_{k=1}^{K} p(r_k^c \mid r_1^1, \dots, r_{k-1}^c, \psi(t))
\end{equation}
where $\psi(t)$ represents the text conditioning.

\medskip

\noindent \textbf{Residual Feature Accumulation.} 
At each scale $k$, the transformer predicts a distribution over tokens which is then quantized to produce the discrete token map $r_k$. This token map represents the residual information needed to refine the video at the current resolution. The feature map $f_k$ is updated by integrating the interpolated results of the current token map $r_k$ with the feature state from the previous scale $f_{k-1}$ via element-wise addition:
\begin{equation}
f_k = \text{Interpolate}(r_k, (h_K, w_K)) + f_{k-1}
\end{equation}
This recursive relationship ensures that $f_k$ encapsulates the integrated information from all preceding scales, providing a stable foundation for further refinement. To further enhance quality, certain scales repeat this process multiple times at the same resolution before advancing to the next scale. The accumulated feature map is downsampled to the next scale resolution $(h_{k+1}, w_{k+1})$:
\begin{equation}
r_{k+1} = \text{Interpolate}(f_k, (h_{k+1}, w_{k+1}))
\end{equation}
The resulting token map $r_{k+1}$ then functions as the primary input for the transformer block at the subsequent scale $k+1$.

\subsection{Motivation} % Observation
This work presents an acceleration framework for STAR-based video synthesis by exploiting the scale-wise information redundancy inherent in hierarchical video feature maps. Specifically, we characterize the model through the spectral convergence of information across resolutions, the spatiotemporal duality of feature maps, and the structural justification for prioritizing token pruning over merging in the context of cumulative refinement.

\begin{figure}[t]
    \centering
    \includegraphics[width=\textwidth]{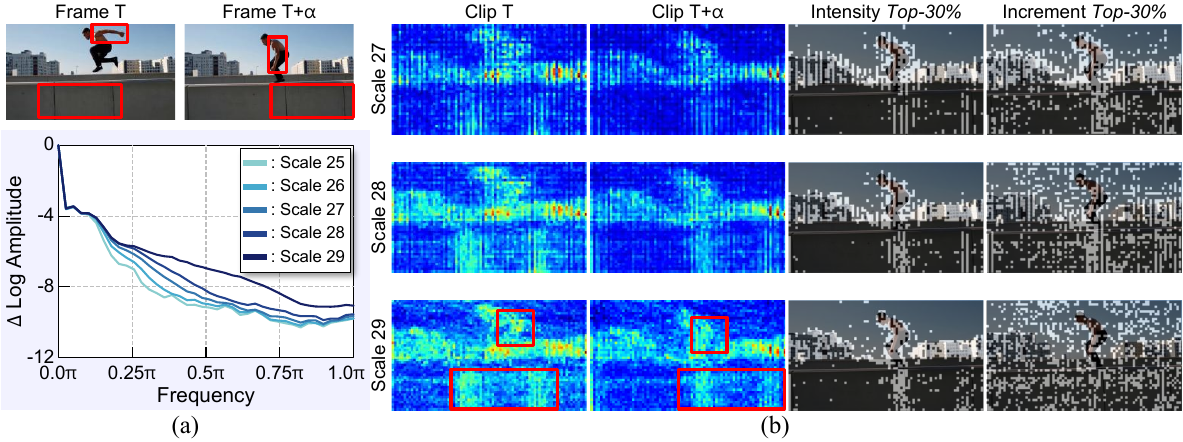}
    \caption{(a) Scale-wise Fourier-based spectral analysis demonstrates early low-frequency convergence alongside sustained high-frequency updates. (b) High-frequency energy maps show intense localization. Comparing high-intensity locations with large incremental updates reveals significant opportunities for token pruning in converged regions.
    }
    \label{method_1}
    \vspace{-0.2in}
    % \vspace{-0.06in}
\end{figure}

\noindent \textbf{\textit {Observation 1: Spectral Convergence of Video Feature Maps.}} \label{method:fourier}
Scale-specific information density in the STAR-based video generation pipeline reveals non-uniform distribution across tokens and scales. As demonstrated by the Fourier-based spectral analysis in Fig.~\ref{method_1}(a), low-frequency components representing the global structural sketch reach convergence early, exhibiting minimal variance across subsequent late scales. In contrast, high-frequency details, such as moving objects and sharp boundaries, are continuously updated until the final scale. This phenomenon is further substantiated in Fig.~\ref{method_1}(b), which presents heatmaps generated by applying an Inverse Fourier Transform (IFT) exclusive to the high-frequency coefficients. These visualizations confirm that high-frequency energy is intensely localized on salient regions, such as human silhouettes and critical background textures. In contrast, the residual increments between successive scales exhibit a scattered, patternless distribution across the entire region. This discrepancy indicates that numerical updates in low-frequency areas fail to emerge as meaningful high-frequency refinements, instead representing redundant computations. Consequently, this localized stability offers a significant opportunity to prune converged regions, concentrating computation on high-frequency areas.

\medskip

\noindent \textbf{\textit {Observation 2: Spatiotemporal Duality of Video Feature Maps.}}
%Unlike static images, tokens in the STAR-based video generation framework exhibit a dual nature by simultaneously capturing spatial textures and temporal dynamics. Within the accumulated video feature map, each token serves as a carrier for both motion evolution and the decomposed visual details across individual frames.
Unlike static images, tokens in the STAR-based framework embody a fundamental spatiotemporal duality, acting as integrated carriers within the accumulated feature map to simultaneously represent motion evolution and decomposed visual textures. As illustrated in Fig.~\ref{method_1}(b), high-frequency tokens are not limited to static spatial regions but are densely distributed along motion trajectories—including both the dynamic paths of subjects and the shifting boundaries of the structure. This signifies that the high-frequency energy effectively encodes the temporal continuity required to decode fluid motion and evolving scene geometry. Consequently, redundancy in STAR is fundamentally distinct from that in static image models, necessitating a pruning criterion that transcends simple spatial importance. Integrating both spatial and temporal characteristics is therefore essential to identify tokens that require active updates while bypassing those that have already reached spatiotemporal convergence.

\begin{figure}[t]
    \centering
    \includegraphics[width=\textwidth]{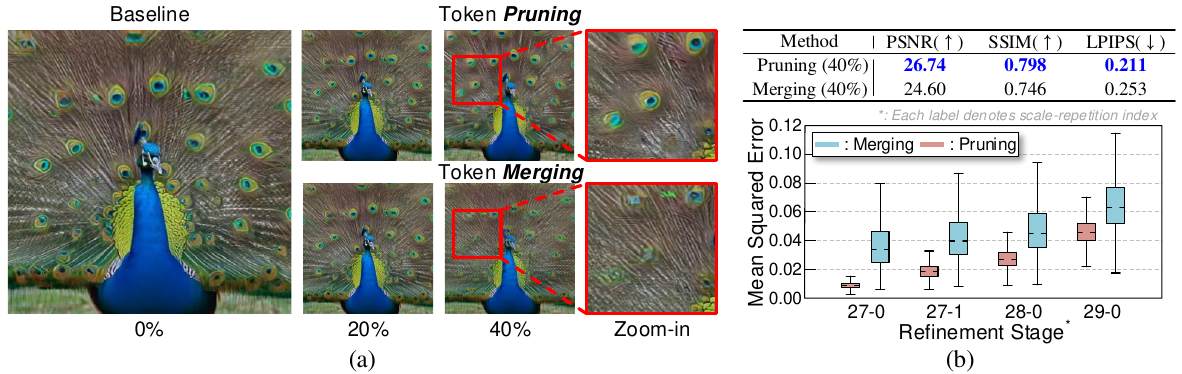}
    \vspace{-0.2in}
    \caption{Comparison of token pruning and merging. (a) Token merging at 40\% obliterates high-frequency textures, while pruning preserves fine details even at high compression. (b) Merging exhibits higher average MSE and expanding variance across resolution scales, whereas pruning maintains tighter error bounds.}
    \label{method_2}
    \vspace{-0.2in}
    % \vspace{-0.06in}
\end{figure}

\medskip

\noindent \textbf{\textit {Observation 3: Structural Compatibility of Pruning over Merging.}} 
As the STAR-based framework accumulates discrete features through quantization, conventional token merging faces fundamental structural limitations. Averaging similar tokens distorts discrete latent representations, triggering an error feedback loop that spreads spatially as resolution increases. This structural incompatibility is empirically evident in Fig.~\ref{method_2}(a), where a 40\% merging ratio obliterates high-frequency patterns. In contrast, pruning---which preserves the spatial integrity of randomly selected locations from the initial scale---effectively maintains fine textures. Quantitative analysis in Fig.~\ref{method_2}(b) further shows that while both methods exhibit rising Mean Squared Error (MSE) with resolution, merging suffers from a significantly higher average MSE and rapidly expanding variance. Conversely, pruning maintains a much tighter variance even at final scales, demonstrating its superior ability to suppress error propagation. By preserving the spatial integrity of selected tokens, pruning ensures a consistent path for residual updates, preventing the representation drift inherent in merging.

\subsection{Spatiotemporal Similarity-based Token Pruning and Partial Update} \label{method:sttp}

\begin{figure}[t]
    \centering
    \includegraphics[width=\textwidth]{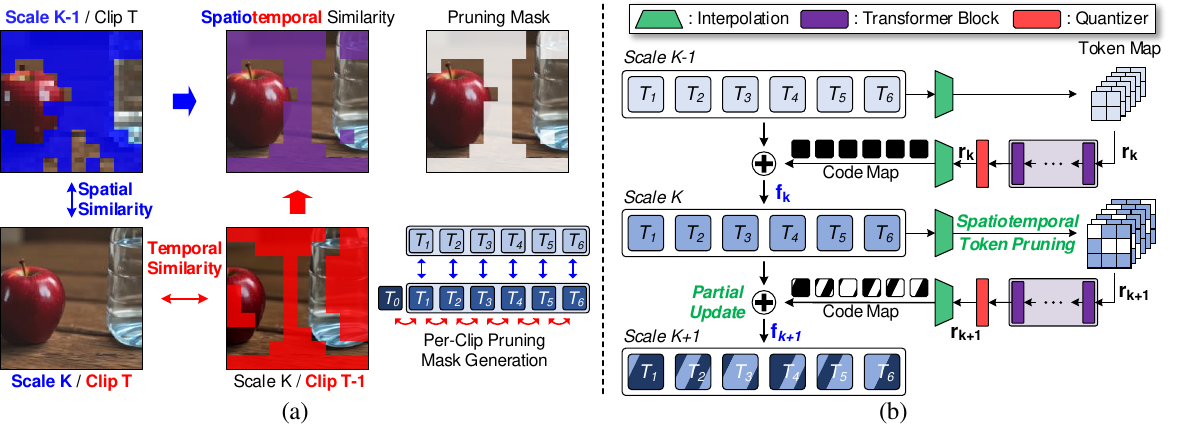}
    \caption{Overview of Spatiotemporal Token Pruning (STTP) and Partial Update (PU). (a) Spatial similarity and temporal similarity are fused into a unified spatiotemporal score to generate a per-clip pruning mask. (b) Only high-priority tokens are passed through the transformer and quantize at each scale, while unselected regions are filled with zeros via Partial Update to preserve the integrity of the cumulative feature map.}
    \label{method_3}
    \vspace{-0.2in}
    % \vspace{-0.06in}
\end{figure}

Building on the observed spectral convergence and spatiotemporal duality of video feature maps, we implement spatiotemporal similarity-based token pruning and a corresponding partial update mechanism to minimize redundant computations within the STAR transformer backbone. This process ensures that only the most informative regions of the video feature map are refined while preserving the structural integrity of already converged regions.

\medskip

\begin{figure}[t]
    \centering
    \includegraphics[width=\textwidth]{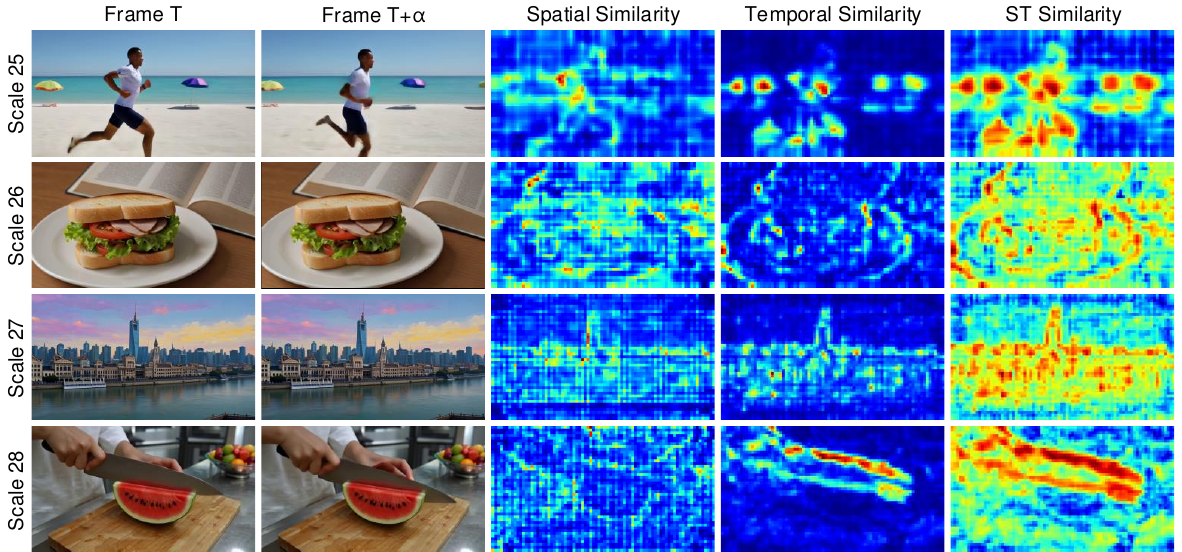}
    \caption{Visualization of similarity maps across scales and frames. Warmer colors (\textcolor{red}{red}) indicate \textcolor{red}{low similarity}, corresponding to dynamic or high-frequency regions requiring update, while cooler colors (\textcolor{blue}{blue}) indicate \textcolor{blue}{high similarity} in converged, static regions.}
    \label{method_4}
    \vspace{-0.1in}
    % \vspace{-0.06in}
\end{figure}

\noindent \textbf{Spatiotemporal Token Pruning (STTP).}
To identify tokens that exhibit persistent high-frequency energy and require continuous updates, we introduce a pruning methodology based on two distinct similarity metrics derived directly from the video feature maps. These metrics allow us to pinpoint regions of high informational gain with minimal computational overhead. Our strategy employs the inherent structural and motion-related information within the feature maps to guide the pruning process:
\begin{itemize}
\item \textbf{Spatial Similarity:} To detect spatial refinements across scales, we compute the cosine similarity between feature maps of the previous and current scales:
\begin{equation}
% \text{Sim}_{\text{spatial}, t} = \frac{f_{t, k-1} \cdot f_{t, k}}{\|f_{t, k-1}\| \|f_{t, k}\|}
s_{t, k}^{\text{spatial}} = \frac{f_{t, k-1} \cdot f_{t, k}}{\|f_{t, k-1}\| \|f_{t, k}\|}, \quad t = 1, \dots, T
\end{equation}
As illustrated in Fig.~\ref{method_4}, lower similarity scores are intensely localized on salient objects and detailed textures, indicating that these high-frequency regions undergo substantive updates to enhance visual fidelity. This allow us to identify non-converged tokens requiring further refinement while preserving already converged areas.
 % To detect intricate spatial refinements across resolutions, we compute the cosine similarity between corresponding feature maps from the previous and current scales. As illustrated in Fig.~\ref{method_4}, lower similarity scores are intensely localized on salient objects and detailed textures, indicating that these high-frequency regions are actively updated to enhance visual fidelity. This spatial divergence allows our model to identify non-converged tokens that require further computational refinement while preserving the integrity of stable, already converged areas.
\item \textbf{Temporal Similarity:}
To exploit temporal continuity within the video feature map, we compute the cosine similarity between the feature maps of clip $t$ and its predecessor $t-1$:
\begin{equation}
% \text{Sim}_{\text{temporal}} = \frac{f_k^{T-1} \cdot f_k^{T}}{\|f_k^{T-1}\| \|f_k^{T}\|}
s_{t, k}^{\text{temporal}} = \frac{f_{t-1, k} \cdot f_{t, k}}{\|f_{t-1, k}\| \|f_{t, k}\|}, \quad t = 1, \dots, T
\end{equation}
As illustrated in Fig.~\ref{method_4}, lower similarity scores are concentrated along motion trajectories, identifying regions that require temporal information to represent fluid motion. For the initial clip $f_{1, k}$, temporal importance is determined by comparing with the last scale feature map of the first clip $f_{0, K}$.
\end{itemize}
 % To exploit temporal continuity within the video feature map, we compute the cosine similarity between the clip $T$ and its predecessor $T-1$, leveraging the causal 3D convolutional structure of the decoder. As illustrated in Fig.~\ref{method_4}, lower similarity scores are concentrated along motion trajectories, identifying where the decoder requires differentiated temporal information to represent fluid motion. For the initial clip of a sequence, temporal importance is determined by comparing its feature map with the already generated feature map of the first clip.

% \noindent \textbf{Metric Fusion via Spatiotemporal Similarity.} To robustly integrate these indicators into a unified score, we transform the cosine similarity values (ranging from $[-1, 1]$) into a positive importance scale where lower similarity yields a higher score. The fusion is defined by a squared L2-norm approach:
% \begin{equation} \text{Score}_{ST} = (1 - \text{Sim}_{spatial})^2 + (1 - \text{Sim}_{temporal})^2 \end{equation}
% By employing a quadratic summation, we create a parabolic score landscape that naturally emphasizes tokens where either spatial or temporal similarity is significantly low. This design ensures that the relative importance of both cues is reflected continuously while providing a steeper gradient for high-priority regions. This quadratic formulation prevents abrupt fluctuations in the final mask, ensuring a stable selection for the binary pruning mask.

% \noindent \textbf{Metric Fusion via Spatiotemporal Similarity.}
\noindent \textbf{Joint Metric Fusion via $\ell_p$-norm Dissimilarity.} To robustly integrate these indicators into a unified score, we transform the cosine similarity values into a dissimilarity-based importance metric; lower similarity signifies a higher refinement priority. The fusion is defined as a $\ell_p$-norm over the two dissimilarity terms:
\begin{equation}
    \text{Score}_{ST} = \left[ (1 - \text{s}_{\text{spatial}})^p + (1 - \text{s}_{\text{temporal}})^p \right]^{1/p}
\end{equation}
where $p$ modulates the sensitivity to high-dissimilarity regions, higher values of $p$ effectively prioritize tokens that exhibit significant change in \textit{either} the spatial or temporal domain, acting as a soft-maximum operator. Conversely, lower values of $p$ provide a more balanced integration of both terms. We empirically validate the choice of $p$ through ablation studies in \S~\ref{exp:ablation}, and set $p=2$ for all experiments.

\subsubsection{Implementation of Partial Update (PU).}
To ensure the pruning mask $M \in \{0, 1\}^{h_K \times w_K}$ is compatible with the hierarchical structure of VAR, we downsample the generated pruning mask to match the current token map dimensions $(h_k, w_k)$ via interpolation. The pruning process for the input token map $r_k$ is defined as follows:\begin{equation}\hat{r}_k = r_k \odot \text{Interpolate}(M, (h_k, w_k))\end{equation}The transformer backbone subsequently processes only the selected tokens $\hat{r}_k$, significantly reducing the computational overhead. After the transformer iterations and quantization are complete, the resulting residual code map is upsampled back to the video feature map resolution $(h_K, w_K)$. To maintain the structural integrity of the cumulative VAR process, we implement a partial update strategy. The unselected regions in the expanded residual map are explicitly filled with zeros using the original pruning mask $M$ before being integrated into the previous video feature map $f_{k-1}$. The partial update for the video feature map $f_k$ is formulated as:\begin{equation}f_k = \text{Interpolate}(\hat{r}_k, (h_K, w_K)) \odot M + f_{k-1} \end{equation}
This mechanism prevents uncomputed or stale noise from polluting converged regions, ensuring that computational resources are strictly concentrated on the most dynamic and detailed portions of the video.
\section{Experiments}

\subsection{Experimental Setup}

\textbf{Models and Baselines.} We evaluate the proposed \methodname using InfinityStar~\cite{infinitystar} as the backbone. To ensure a comprehensive comparison, we implemented representative training-free token reduction baselines: SparseVAR~\cite{sparsevar}, FastVAR~\cite{fastvar}, and ToMe~\cite{tome}. Since these methods were originally designed for text-to-image synthesis or non-autoregressive architectures, we extended them to accommodate the temporal dimension by applying their respective reduction mechanisms across the spatiotemporal pyramid. This ensures a fair and competitive evaluation against established image-based acceleration strategies within the video domain. For all experiments, we maintain the default hyperparameters of the backbone model to ensure consistency.

\medskip

\noindent \textbf{Metrics.} We assess our framework across two primary dimensions: inference efficiency and generation quality.

\begin{itemize}
    \item \textbf{Inference Efficiency:} We measure the practical acceleration of our method on a NVIDIA H100 GPU (80GB). Efficiency metrics include end-to-end inference latency across the text encoder and VAE decoder, and the achieved speedup.
    \item \textbf{Generation Quality:} We utilize the VBench~\cite{vbench} suite, evaluating 5-second videos (81 frames) at 720p resolution across all 16 dimensions using the official VBench prompt set. To ensure a fair comparison, we employ the same rewritten prompts used in the evaluation of InfinityStar. Additionally, we report Peak Signal-to-Noise Ratio (PSNR),  Structural Similarity Index Measure (SSIM), and Learned Perceptual Image Patch Similarity (LPIPS)~\cite{lpips} to quantify reconstruction fidelity and perceptual similarity. These metrics are computed on 10 videos sampled per VBench dimension for each task (T2V, I2V, and V2V).
\end{itemize}

\noindent \textbf{Implementations.} For the spatiotemporal similarity-based pruning, we apply STTP to the high-resolution refinement stages, which account for the vast majority of the total inference latency. Specifically, we focus on the final four scales to exploit cross-scale and intra-scale redundancies during the synthesis process, alleviating the computational burden while preserving the established structural layouts. For scales with multiple iterations at the same resolution, pruning is applied to early iterations, reserving the final iteration for full-token refinement.

\subsection{Comparative Analysis on Text-to-Video Generation}

\begin{table}[t]
    \centering
    \caption{Quantitative and efficiency benchmarking for 720p, 5s (81 frames) Text-to-Video generation. We compare \methodname with baselines across quality metrics and inference latency; \textbf{best results} are shown in bold and \underline{second-best results} are underlined; $\dagger$ denotes results using prompt rewriting.}
    \includegraphics[width=\textwidth]{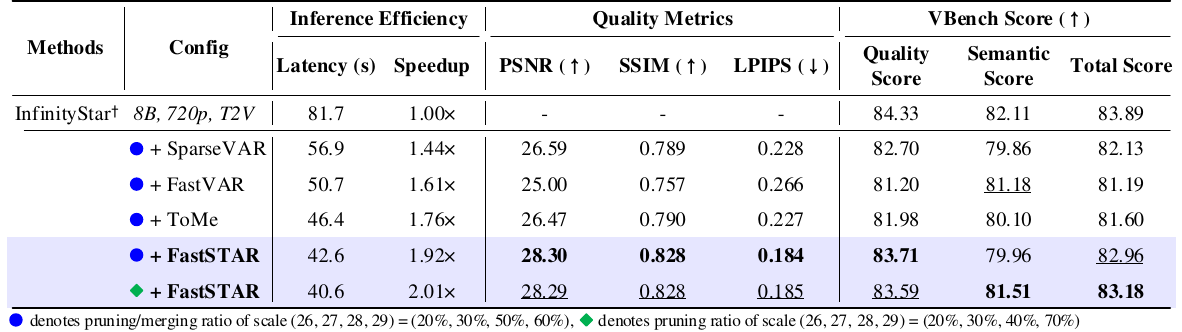}
    \label{exp_1}
    \vspace{-0.1in}
    % \vspace{-0.06in}
\end{table}

\newcommand{\blkhl}[1]{\colorbox{black!20}{#1}}
\newcommand{\whthl}[1]{\fcolorbox{white}{white}{#1}}

\begin{figure}[t]
    \centering
    \vspace{-0.1in}
    \includegraphics[width=\textwidth]{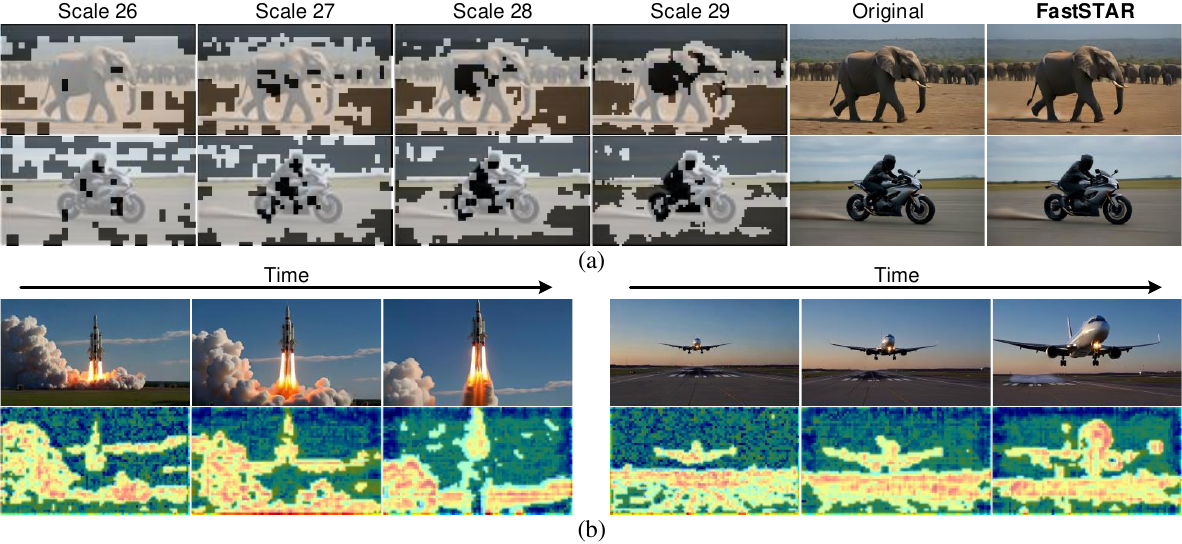}
    \caption{Qualitative results.
    (a) Pruning masks across scales and resulting outputs.
    (b) Frame-level results at the last scale with spatiotemporal similarity heatmaps and pruning masks.
    \blkhl{Black} denotes pruned regions and \whthl{white} denotes computed regions.}
    \label{exp_2}
    \vspace{-0.2in}
\end{figure}

\noindent \textbf{Quantitative Comparison.} Table~\ref{exp_1} compares the inference efficiency and generation quality of various acceleration schemes applied to the InfinityStar baseline for T2V. Although the base InfinityStar model offers high-fidelity video synthesis, its original inference time of 81.7s imposes a significant computational burden for 720p generation. While existing reduction methods like SparseVAR, FastVAR, and ToMe yield speedups of $1.44\times$, $1.61\times$, and $1.76\times$, respectively, they suffer substantial declines in both VBench scores and Quality Metrics, including PSNR, SSIM, and LPIPS. This degradation stems from their inherent limitations in handling the hierarchical and temporal complexities of STAR. In contrast, \methodname maintains high generation quality across different pruning configurations. Specifically, for two configurations with similar pruning ratios—(20\%, 30\%, 50\%, 60\%) and (20\%, 30\%, 40\%, 70\%)—our method achieves acceleration up to $2.01\times$ with a negligible VBench score reduction of only 0.85\% from the baseline. Furthermore, our method significantly outperforms all other baselines in quality metrics, achieving the highest PSNR of 28.29. \label{exp:vbench}

%(28.29) and SSIM (0.828) while maintaining the lowest LPIPS (0.185) among the accelerated models.
 % Due to space constraints, we provide a comprehensive breakdown of the VBench scores across all 16 dimensions in Appendix~\ref{app:vbench_details}.
 
\medskip

\noindent \textbf{Qualitative Comparison.} Fig.~\ref{exp_2} provides a qualitative visualization of how our STTP accurately identifies redundant regions. As illustrated in Fig.~\ref{exp_2}(a), as the spatial scale increases, \methodname effectively captures areas where further updates are no longer required, spanning both static backgrounds and converged regions within dynamic objects. Furthermore, Fig.~\ref{exp_2}(b) displays generated video frames overlaid with our fused similarity score heatmaps and the resulting pruning masks. Because temporal characteristics are natively fused into the importance scores, the heatmaps clearly demonstrate that \methodname preserves critical motion trajectories while aggressively pruning spatiotemporally redundant tokens, ensuring fluid motion and structural integrity.

\subsection{Comparative Analysis on Image-to-Video Synthesis}

\begin{table}[t]
    \centering
    \caption{Comparison with state-of-the-art token reduction methods on 720p, 5s (81 frames) Image-to-Video generation; \textbf{best results} are shown in bold and \underline{second-best results} are underlined.}
    \vspace{-0.1in}
    \includegraphics[width=\textwidth]{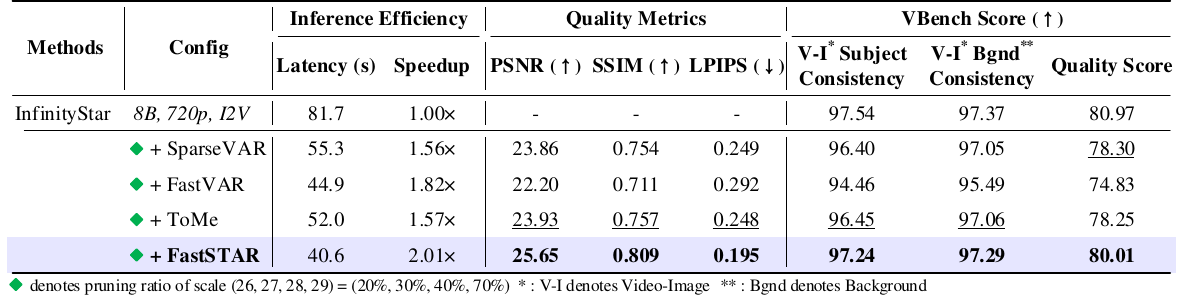}
    \label{exp_3}
    \vspace{-0.2in}
    % \vspace{-0.06in}
\end{table}
\begin{figure}[t]
    \centering
    \includegraphics[width=\textwidth]{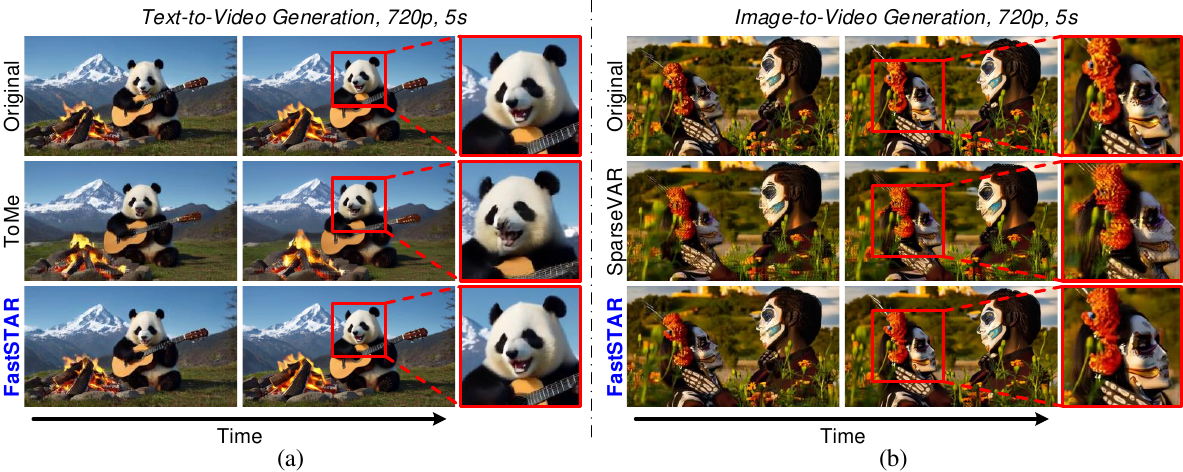}
    \caption{Qualitative comparison of generation quality. (a) T2V. (b) I2V.}
    \label{exp_6}
    \vspace{-0.2in}
    % \vspace{-0.06in}
\end{figure}

\noindent \textbf{Quantitative Comparison.} We evaluate the performance of \methodname against established token reduction baselines, including SparseVAR, FastVAR, and ToMe, as shown in Table~\ref{exp_3}. For a fair comparison, we align the pruning and merging ratios at $(20\%, 30\%, 40\%, 70\%)$ across the final scales of the 720p I2V task. While all methods achieve speedups, \methodname significantly outperforms the baselines in maintaining quality metrics, reporting superior scores in PSNR of 25.65.
%), SSIM (0.809), and LPIPS (0.195).

\medskip

\noindent \textbf{Qualitative Comparison.} Fig.~\ref{exp_6} compares the generation quality across different tasks and acceleration methods. As illustrated in the T2V task in Fig.~\ref{exp_6}(a), \methodname maintains nearly identical visual quality to the original baseline, preserving intricate textures and structural details that are significantly degraded in ToMe. Similarly, Fig.~\ref{exp_6}(b) demonstrates that in the I2V task, \methodname effectively avoids the severe artifacts and loss of semantic clarity seen in SparseVAR. By accurately identifying redundant regions through spatiotemporal awareness, \methodname ensures superior structural integrity and visual consistency, even under the aggressive pruning ratios required for $2.01\times$ acceleration.

\subsection{Robustness Across Various Synthesis Tasks}
To demonstrate the versatility of our framework, we apply \methodname to various synthesis tasks at 480p resolution, including T2V, I2V, and V2V. As summarized in Table~\ref{exp_4}, our method consistently achieves acceleration across all these tasks while maintaining high reconstruction fidelity. Notably, the acceleration gain at $480$p is slightly lower than at $720$p due to the reduced scale count and smaller token map dimensions. These results confirm robustness across different conditioning signals and resolutions.
%Furthermore, \methodname efficiently processes sequences up to 81 frames without resolution-specific degradation.

\begin{table}[t]
    \centering
    \caption{Performance across 480p STAR-based synthesis tasks (T2V, I2V, V2V).}
    \vspace{-0.1in}
    \includegraphics[width=\textwidth]{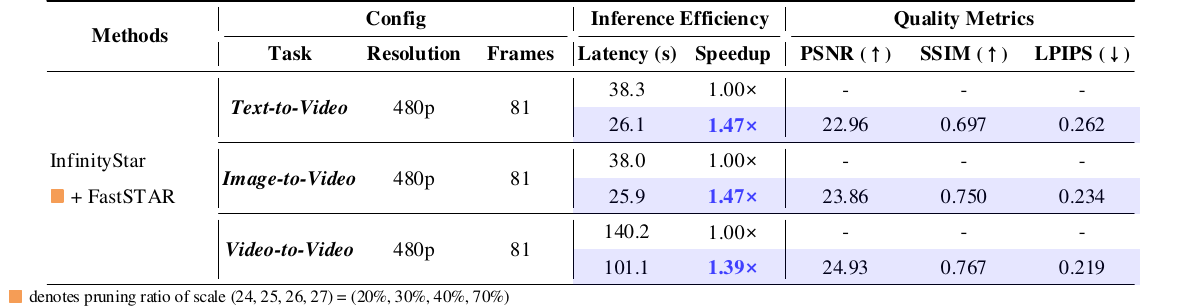}
    \label{exp_4}
    \vspace{-0.2in}
    % \vspace{-0.06in}
\end{table}

\subsection{Ablation Studies} \label{exp:ablation}

\begin{figure}[t]
    \centering
    \includegraphics[width=\textwidth]{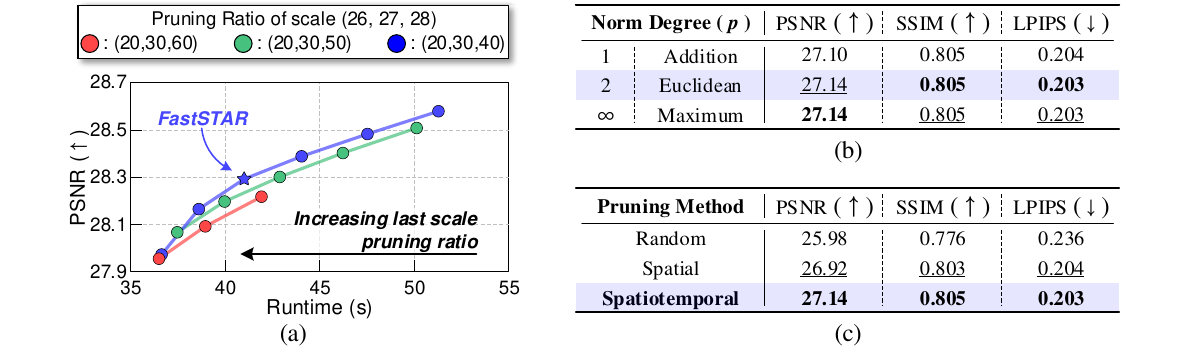}
    \caption{Ablation studies of \methodname.
    (a) Effect of pruning ratios on runtime and PSNR.
    (b) Sensitivity to norm degree $p$ in score fusion.
    (c) Comparison of pruning methods.
    \textbf{Best results} are shown in bold and \underline{second-best results} are underlined.}
    \label{exp_5}
    \vspace{-0.2in}
    % \vspace{-0.06in}
\end{figure}

\noindent \textbf{Different Pruning Ratios.} The pruning ratio dictates the balance between computational efficiency and generation quality. As shown in Fig.~\ref{exp_5}(a), increasing the pruning ratio in the final scales results in a direct reduction in runtime, albeit with an expected trade-off in reconstruction fidelity. Through our analysis across scales 26, 27, and 28, we observe that configurations of $(20\%, 30\%, 40\%)$ maintain a stable quality-to-speed slope. Specifically, for the final scale (scale 29), we find that quality remains robust up to a $70\%$ pruning ratio; however, exceeding this threshold leads to severe quality degradation. Consequently, we adopt the $(20\%, 30\%, 40\%, 70\%)$ configuration as our primary setting to maximize speedup while preserving structural integrity.

\medskip

\noindent \textbf{Norm Degree.} To analyze the sensitivity of our spatiotemporal score fusion to the choice of norm degree $p$, we compare $p \in \{1, 2, \infty\}$ in Fig.~\ref{exp_5}(b). As shown, while all choices yield competitive results, $p=2$ achieves the best perceptual quality, recording the highest SSIM and lowest LPIPS. This aligns with its role as a balanced integrator between the maximum behavior of $p=\infty$ and the uniform weighting of $p=1$, and is therefore adopted as our default configuration.

\medskip

\noindent \textbf{Pruning Method.} To justify the necessity of our spatiotemporal approach, we compare its performance against random and purely spatial-based pruning strategies in Fig.~\ref{exp_5}(c). Random pruning fails to preserve critical information, resulting in poor quality across all dimensions. While spatial-only pruning offers an improvement, it lacks the temporal context required to maintain fluid motion and consistency. \methodname surpasses both baselines across the entire spectrum of metrics, establishing that integrating temporal dynamics into the pruning method is essential for high-fidelity video synthesis.

\definecolor{MyLightGreen}{RGB}{0,176,80}

\section{Conclusion}

In this work, we presented \methodname, a training-free acceleration framework specifically designed for high-quality video generation within the Spacetime AutoRegressive modeling (STAR). Our core innovation, Spatiotemporal Token Pruning (STTP), addresses the "token explosion" at final refinement scales by integrating spatial structural convergence and temporal motion trajectories to identify high-priority tokens. Unlike conventional merging strategies that distort discrete feature distributions, our pruning-over-merging approach, combined with a Partial Update (PU) mechanism, effectively prevents error propagation while bypassing redundant computations.

Experimental results on the InfinityStar model demonstrate that \methodname achieves a significant end-to-end speedup of up to 2.01$\times$ on a single NVIDIA H100 GPU for 720p video synthesis. Remarkably, this acceleration is achieved with a negligible VBench score degradation of less than 1\%, consistently delivering robust performance across various generation modes including T2V, I2V, and V2V. By defining a new Pareto frontier for efficient video synthesis, \methodname provides a practical and scalable solution for high-resolution autoregressive modeling in real-world applications.

\newpage

% ---- Bibliography ----
%
% BibTeX users should specify bibliography style 'splncs04'.
% References will then be sorted and formatted in the correct style.
%
\bibliographystyle{splncs04}
\bibliography{main}

@String(CVPR  = {IEEE Conf. Comput. Vis. Pattern Recog.})

@String(ICCV  = {Int. Conf. Comput. Vis.})

@String(AAAI  = {AAAI})

@String(CVPR  = {CVPR})

@String(ICCV  = {ICCV})

@inproceedings{var,
    author = {Tian, Keyu and Jiang, Yi and Yuan, Zehuan and Peng, Bingyue and Wang, Liwei},
    booktitle = {Advances in Neural Information Processing Systems},
    doi = {10.52202/079017-2694},
    editor = {A. Globerson and L. Mackey and D. Belgrave and A. Fan and U. Paquet and J. Tomczak and C. Zhang},
    pages = {84839--84865},
    publisher = {Curran Associates, Inc.},
    title = {Visual Autoregressive Modeling: Scalable Image Generation via Next-Scale Prediction},
    url = {https://proceedings.neurips.cc/paper_files/paper/2024/file/9a24e284b187f662681440ba15c416fb-Paper-Conference.pdf},
    volume = {37},
    year = {2024}
}

@inproceedings{infinity,
    author    = {Han, Jian and Liu, Jinlai and Jiang, Yi and Yan, Bin and Zhang, Yuqi and Yuan, Zehuan and Peng, Bingyue and Liu, Xiaobing},
    title     = {Infinity: Scaling Bitwise AutoRegressive Modeling for High-Resolution Image Synthesis},
    booktitle = {Proceedings of the IEEE/CVF Conference on Computer Vision and Pattern Recognition (CVPR)},
    month     = {June},
    year      = {2025},
    pages     = {15733-15744}
}

@misc{infinitystar,
      title={InfinityStar: Unified Spacetime AutoRegressive Modeling for Visual Generation}, 
      author={Jinlai Liu and Jian Han and Bin Yan and Hui Wu and Fengda Zhu and Xing Wang and Yi Jiang and Bingyue Peng and Zehuan Yuan},
      year={2025},
      eprint={2511.04675},
      archivePrefix={arXiv},
      primaryClass={cs.CV},
      url={https://arxiv.org/abs/2511.04675}, 
}

@misc{videoar,
      title={VideoAR: Autoregressive Video Generation via Next-Frame \& Scale Prediction}, 
      author={Longbin Ji and Xiaoxiong Liu and Junyuan Shang and Shuohuan Wang and Yu Sun and Hua Wu and Haifeng Wang},
      year={2026},
      eprint={2601.05966},
      archivePrefix={arXiv},
      primaryClass={cs.CV},
      url={https://arxiv.org/abs/2601.05966}, 
}

@misc{nxtframe,
      title={Autoregressive Video Generation beyond Next Frames Prediction}, 
      author={Sucheng Ren and Chen Chen and Zhenbang Wang and Liangchen Song and Xiangxin Zhu and Alan Yuille and Yinfei Yang and Jiasen Lu},
      year={2025},
      eprint={2509.24081},
      archivePrefix={arXiv},
      primaryClass={cs.CV},
      url={https://arxiv.org/abs/2509.24081}, 
}

@misc{emu3,
      title={Emu3: Next-Token Prediction is All You Need}, 
      author={Xinlong Wang and Xiaosong Zhang and Zhengxiong Luo and Quan Sun and Yufeng Cui and Jinsheng Wang and Fan Zhang and Yueze Wang and Zhen Li and Qiying Yu and Yingli Zhao and Yulong Ao and Xuebin Min and Tao Li and Boya Wu and Bo Zhao and Bowen Zhang and Liangdong Wang and Guang Liu and Zheqi He and Xi Yang and Jingjing Liu and Yonghua Lin and Tiejun Huang and Zhongyuan Wang},
      year={2024},
      eprint={2409.18869},
      archivePrefix={arXiv},
      primaryClass={cs.CV},
      url={https://arxiv.org/abs/2409.18869}, 
}

@misc{nova,
      title={Autoregressive Video Generation without Vector Quantization}, 
      author={Haoge Deng and Ting Pan and Haiwen Diao and Zhengxiong Luo and Yufeng Cui and Huchuan Lu and Shiguang Shan and Yonggang Qi and Xinlong Wang},
      year={2025},
      eprint={2412.14169},
      archivePrefix={arXiv},
      primaryClass={cs.CV},
      url={https://arxiv.org/abs/2412.14169}, 
}

@misc{lumos,
      title={Lumos-1: On Autoregressive Video Generation from a Unified Model Perspective}, 
      author={Hangjie Yuan and Weihua Chen and Jun Cen and Hu Yu and Jingyun Liang and Shuning Chang and Zhihui Lin and Tao Feng and Pengwei Liu and Jiazheng Xing and Hao Luo and Jiasheng Tang and Fan Wang and Yi Yang},
      year={2025},
      eprint={2507.08801},
      archivePrefix={arXiv},
      primaryClass={cs.CV},
      url={https://arxiv.org/abs/2507.08801}, 
}

@misc{videogpt,
      title={VideoGPT: Video Generation using VQ-VAE and Transformers}, 
      author={Wilson Yan and Yunzhi Zhang and Pieter Abbeel and Aravind Srinivas},
      year={2021},
      eprint={2104.10157},
      archivePrefix={arXiv},
      primaryClass={cs.CV},
      url={https://arxiv.org/abs/2104.10157}, 
}

@inproceedings{magvit,
  title={MAGVIT: Masked Generative Video Transformer},
  author={Yu, Lijun and Cheng, Yong and Sohn, Kihyuk and Lezama, Jose and Zhang, Han and Chang, Huiwen and Hauptmann, Alexander G and Yang, Ming-Hsuan and Hao, Yuan and Essa, Irfan and others},
  booktitle={2023 IEEE/CVF Conference on Computer Vision and Pattern Recognition (CVPR)},
  pages={10459--10469},
  year={2023},
  organization={IEEE Computer Society}
}

@inproceedings{fastvar,
    author    = {Guo, Hang and Li, Yawei and Zhang, Taolin and Wang, Jiangshan and Dai, Tao and Xia, Shu-Tao and Benini, Luca},
    title     = {FastVAR: Linear Visual Autoregressive Modeling via Cached Token Pruning},
    booktitle = {Proceedings of the IEEE/CVF International Conference on Computer Vision (ICCV)},
    month     = {October},
    year      = {2025},
    pages     = {19011-19021}
}

@misc{tome,
      title={Token Merging: Your ViT But Faster}, 
      author={Daniel Bolya and Cheng-Yang Fu and Xiaoliang Dai and Peizhao Zhang and Christoph Feichtenhofer and Judy Hoffman},
      year={2023},
      eprint={2210.09461},
      archivePrefix={arXiv},
      primaryClass={cs.CV},
      url={https://arxiv.org/abs/2210.09461}, 
}

@misc{sparsevar,
      title={Frequency-Aware Autoregressive Modeling for Efficient High-Resolution Image Synthesis}, 
      author={Zhuokun Chen and Jugang Fan and Zhuowei Yu and Bohan Zhuang and Mingkui Tan},
      year={2025},
      eprint={2507.20454},
      archivePrefix={arXiv},
      primaryClass={cs.CV},
      url={https://arxiv.org/abs/2507.20454}, 
}

@misc{skipvar,
      title={SkipVAR: Accelerating Visual Autoregressive Modeling via Adaptive Frequency-Aware Skipping}, 
      author={Jiajun Li and Yue Ma and Xinyu Zhang and Qingyan Wei and Songhua Liu and Linfeng Zhang},
      year={2025},
      eprint={2506.08908},
      archivePrefix={arXiv},
      primaryClass={cs.CV},
      url={https://arxiv.org/abs/2506.08908}, 
}

@misc{stagevar,
      title={StageVAR: Stage-Aware Acceleration for Visual Autoregressive Models}, 
      author={Senmao Li and Kai Wang and Salman Khan and Fahad Shahbaz Khan and Jian Yang and Yaxing Wang},
      year={2025},
      eprint={2512.16483},
      archivePrefix={arXiv},
      primaryClass={cs.CV},
      url={https://arxiv.org/abs/2512.16483}, 
}

@misc{scaleKV,
      title={Memory-Efficient Visual Autoregressive Modeling with Scale-Aware KV Cache Compression}, 
      author={Kunjun Li and Zigeng Chen and Cheng-Yen Yang and Jenq-Neng Hwang},
      year={2025},
      eprint={2505.19602},
      archivePrefix={arXiv},
      primaryClass={cs.LG},
      url={https://arxiv.org/abs/2505.19602}, 
}

@InProceedings{code,
    author    = {Chen, Zigeng and Ma, Xinyin and Fang, Gongfan and Wang, Xinchao},
    title     = {Collaborative Decoding Makes Visual Auto-Regressive Modeling Efficient},
    booktitle = {Proceedings of the IEEE/CVF Conference on Computer Vision and Pattern Recognition (CVPR)},
    month     = {June},
    year      = {2025},
    pages     = {23334-23344}
}

@misc{hunyuanvideo,
      title={HunyuanVideo: A Systematic Framework For Large Video Generative Models}, 
      author={Weijie Kong and Qi Tian and Zijian Zhang and Rox Min and Zuozhuo Dai and Jin Zhou and Jiangfeng Xiong and Xin Li and Bo Wu and Jianwei Zhang and Kathrina Wu and Qin Lin and Junkun Yuan and Yanxin Long and Aladdin Wang and Andong Wang and Changlin Li and Duojun Huang and Fang Yang and Hao Tan and Hongmei Wang and Jacob Song and Jiawang Bai and Jianbing Wu and Jinbao Xue and Joey Wang and Kai Wang and Mengyang Liu and Pengyu Li and Shuai Li and Weiyan Wang and Wenqing Yu and Xinchi Deng and Yang Li and Yi Chen and Yutao Cui and Yuanbo Peng and Zhentao Yu and Zhiyu He and Zhiyong Xu and Zixiang Zhou and Zunnan Xu and Yangyu Tao and Qinglin Lu and Songtao Liu and Dax Zhou and Hongfa Wang and Yong Yang and Di Wang and Yuhong Liu and Jie Jiang and Caesar Zhong},
      year={2025},
      eprint={2412.03603},
      archivePrefix={arXiv},
      primaryClass={cs.CV},
      url={https://arxiv.org/abs/2412.03603}, 
}

@misc{cogvideox,
      title={CogVideoX: Text-to-Video Diffusion Models with An Expert Transformer}, 
      author={Zhuoyi Yang and Jiayan Teng and Wendi Zheng and Ming Ding and Shiyu Huang and Jiazheng Xu and Yuanming Yang and Wenyi Hong and Xiaohan Zhang and Guanyu Feng and Da Yin and Yuxuan Zhang and Weihan Wang and Yean Cheng and Bin Xu and Xiaotao Gu and Yuxiao Dong and Jie Tang},
      year={2025},
      eprint={2408.06072},
      archivePrefix={arXiv},
      primaryClass={cs.CV},
      url={https://arxiv.org/abs/2408.06072}, 
}

@misc{stable_diffusion,
      title={High-Resolution Image Synthesis with Latent Diffusion Models}, 
      author={Robin Rombach and Andreas Blattmann and Dominik Lorenz and Patrick Esser and Björn Ommer},
      year={2022},
      eprint={2112.10752},
      archivePrefix={arXiv},
      primaryClass={cs.CV},
      url={https://arxiv.org/abs/2112.10752}, 
}

@misc{pixart_alpha,
      title={PixArt-$\alpha$: Fast Training of Diffusion Transformer for Photorealistic Text-to-Image Synthesis}, 
      author={Junsong Chen and Jincheng Yu and Chongjian Ge and Lewei Yao and Enze Xie and Yue Wu and Zhongdao Wang and James Kwok and Ping Luo and Huchuan Lu and Zhenguo Li},
      year={2023},
      eprint={2310.00426},
      archivePrefix={arXiv},
      primaryClass={cs.CV},
      url={https://arxiv.org/abs/2310.00426}, 
}

@misc{pixart_sigma,
      title={PixArt-$\Sigma$: Weak-to-Strong Training of Diffusion Transformer for 4K Text-to-Image Generation}, 
      author={Junsong Chen and Chongjian Ge and Enze Xie and Yue Wu and Lewei Yao and Xiaozhe Ren and Zhongdao Wang and Ping Luo and Huchuan Lu and Zhenguo Li},
      year={2024},
      eprint={2403.04692},
      archivePrefix={arXiv},
      primaryClass={cs.CV},
      url={https://arxiv.org/abs/2403.04692}, 
}

@misc{pixart_delta,
      title={PIXART-$\delta$: Fast and Controllable Image Generation with Latent Consistency Models}, 
      author={Junsong Chen and Yue Wu and Simian Luo and Enze Xie and Sayak Paul and Ping Luo and Hang Zhao and Zhenguo Li},
      year={2024},
      eprint={2401.05252},
      archivePrefix={arXiv},
      primaryClass={cs.CV},
      url={https://arxiv.org/abs/2401.05252}, 
}

@misc{ddim,
      title={Denoising Diffusion Implicit Models}, 
      author={Jiaming Song and Chenlin Meng and Stefano Ermon},
      year={2022},
      eprint={2010.02502},
      archivePrefix={arXiv},
      primaryClass={cs.LG},
      url={https://arxiv.org/abs/2010.02502}, 
}

@misc{ddpm,
      title={Denoising Diffusion Probabilistic Models}, 
      author={Jonathan Ho and Ajay Jain and Pieter Abbeel},
      year={2020},
      eprint={2006.11239},
      archivePrefix={arXiv},
      primaryClass={cs.LG},
      url={https://arxiv.org/abs/2006.11239}, 
}

@misc{videolcm,
      title={VideoLCM: Video Latent Consistency Model}, 
      author={Xiang Wang and Shiwei Zhang and Han Zhang and Yu Liu and Yingya Zhang and Changxin Gao and Nong Sang},
      year={2023},
      eprint={2312.09109},
      archivePrefix={arXiv},
      primaryClass={cs.CV},
      url={https://arxiv.org/abs/2312.09109}, 
}

@inproceedings{mcm,
 author = {Zhai, Yuanhao and Lin, Kevin and Yang, Zhengyuan and Li, Linjie and Wang, Jianfeng and Lin, Chung-Ching and Doermann, David and Yuan, Junsong and Wang, Lijuan},
 booktitle = {Advances in Neural Information Processing Systems},
 doi = {10.52202/079017-3524},
 editor = {A. Globerson and L. Mackey and D. Belgrave and A. Fan and U. Paquet and J. Tomczak and C. Zhang},
 pages = {111000--111021},
 publisher = {Curran Associates, Inc.},
 title = {Motion Consistency Model: Accelerating Video Diffusion with Disentangled Motion-Appearance Distillation},
 url = {https://proceedings.neurips.cc/paper_files/paper/2024/file/c859b99b5d717c9035e79d43dfd69435-Paper-Conference.pdf},
 volume = {37},
 year = {2024}
}

@misc{animatediff_lightning,
      title={AnimateDiff-Lightning: Cross-Model Diffusion Distillation}, 
      author={Shanchuan Lin and Xiao Yang},
      year={2024},
      eprint={2403.12706},
      archivePrefix={arXiv},
      primaryClass={cs.CV},
      url={https://arxiv.org/abs/2403.12706}, 
}

@misc{fastercache,
      title={FasterCache: Training-Free Video Diffusion Model Acceleration with High Quality}, 
      author={Zhengyao Lv and Chenyang Si and Junhao Song and Zhenyu Yang and Yu Qiao and Ziwei Liu and Kwan-Yee K. Wong},
      year={2025},
      eprint={2410.19355},
      archivePrefix={arXiv},
      primaryClass={cs.CV},
      url={https://arxiv.org/abs/2410.19355}, 
}

@misc{easycache,
      title={Less is Enough: Training-Free Video Diffusion Acceleration via Runtime-Adaptive Caching}, 
      author={Xin Zhou and Dingkang Liang and Kaijin Chen and Tianrui Feng and Xiwu Chen and Hongkai Lin and Yikang Ding and Feiyang Tan and Hengshuang Zhao and Xiang Bai},
      year={2025},
      eprint={2507.02860},
      archivePrefix={arXiv},
      primaryClass={cs.CV},
      url={https://arxiv.org/abs/2507.02860}, 
}

@inproceedings{vidtome,
    author    = {Li, Xirui and Ma, Chao and Yang, Xiaokang and Yang, Ming-Hsuan},
    title     = {VidToMe: Video Token Merging for Zero-Shot Video Editing},
    booktitle = {Proceedings of the IEEE/CVF Conference on Computer Vision and Pattern Recognition (CVPR)},
    month     = {June},
    year      = {2024},
    pages     = {7486-7495}
}

@article{f3pruning, title={F³-Pruning: A Training-Free and Generalized Pruning Strategy towards Faster and Finer Text-to-Video Synthesis}, volume={38}, url={https://ojs.aaai.org/index.php/AAAI/article/view/28300}, DOI={10.1609/aaai.v38i5.28300}, abstractNote={Recently Text-to-Video (T2V) synthesis has undergone a breakthrough by training transformers or diffusion models on large-scale datasets. Nevertheless, inferring such large models incurs huge costs. Previous inference acceleration works either require costly retraining or are model-specific. To address this issue, instead of retraining we explore the inference process of two mainstream T2V models using transformers and diffusion models. The exploration reveals the redundancy in temporal attention modules of both models, which are commonly utilized to establish temporal relations among frames. Consequently, we propose a training-free and generalized pruning strategy called F3-Pruning to prune redundant temporal attention weights. Specifically, when aggregate temporal attention values are ranked below a certain ratio, corresponding weights will be pruned. Extensive experiments on three datasets using a classic transformer-based model CogVideo and a typical diffusion-based model Tune-A-Video verify the effectiveness of F3-Pruning in inference acceleration, quality assurance and broad applicability.}, number={5}, journal={Proceedings of the AAAI Conference on Artificial Intelligence}, author={Su, Sitong and Liu, Jianzhi and Gao, Lianli and Song, Jingkuan}, year={2024}, month={Mar.}, pages={4961-4969} 
}

@inproceedings{dpmsolver,
 author = {Lu, Cheng and Zhou, Yuhao and Bao, Fan and Chen, Jianfei and LI, Chongxuan and Zhu, Jun},
 booktitle = {Advances in Neural Information Processing Systems},
 editor = {S. Koyejo and S. Mohamed and A. Agarwal and D. Belgrave and K. Cho and A. Oh},
 pages = {5775--5787},
 publisher = {Curran Associates, Inc.},
 title = {DPM-Solver: A Fast ODE Solver for Diffusion Probabilistic Model Sampling in Around 10 Steps},
 url = {https://proceedings.neurips.cc/paper_files/paper/2022/file/260a14acce2a89dad36adc8eefe7c59e-Paper-Conference.pdf},
 volume = {35},
 year = {2022}
}

@misc{toca,
      title={Accelerating Diffusion Transformers with Token-wise Feature Caching}, 
      author={Chang Zou and Xuyang Liu and Ting Liu and Siteng Huang and Linfeng Zhang},
      year={2025},
      eprint={2410.05317},
      archivePrefix={arXiv},
      primaryClass={cs.LG},
      url={https://arxiv.org/abs/2410.05317}, 
}

@inproceedings{tomesd,
    author    = {Bolya, Daniel and Hoffman, Judy},
    title     = {Token Merging for Fast Stable Diffusion},
    booktitle = {Proceedings of the IEEE/CVF Conference on Computer Vision and Pattern Recognition (CVPR) Workshops},
    month     = {June},
    year      = {2023},
    pages     = {4599-4603}
}

@inproceedings{sdtm,
    author    = {Fang, Haipeng and Tang, Sheng and Cao, Juan and Zhang, Enshuo and Tang, Fan and Lee, Tong-Yee},
    title     = {Attend to Not Attended: Structure-then-Detail Token Merging for Post-training DiT Acceleration},
    booktitle = {Proceedings of the IEEE/CVF Conference on Computer Vision and Pattern Recognition (CVPR)},
    month     = {June},
    year      = {2025},
    pages     = {18083-18092}
}

@misc{toma,
      title={ToMA: Token Merge with Attention for Diffusion Models}, 
      author={Wenbo Lu and Shaoyi Zheng and Yuxuan Xia and Shengjie Wang},
      year={2025},
      eprint={2509.10918},
      archivePrefix={arXiv},
      primaryClass={cs.LG},
      url={https://arxiv.org/abs/2509.10918}, 
}

@inproceedings{asymrnr,
  title={AsymRnR: Video Diffusion Transformers Acceleration with Asymmetric Reduction and Restoration},
  author={Sun, Wenhao and Tu, Rong-Cheng and Liao, Jingyi and Jin, Zhao and Tao, Dacheng},
  booktitle={International Conference on Machine Learning},
  pages={57694--57711},
  year={2025},
  organization={PMLR}
}

@misc{diagd,
      title={Fast Autoregressive Video Generation with Diagonal Decoding}, 
      author={Yang Ye and Junliang Guo and Haoyu Wu and Tianyu He and Tim Pearce and Tabish Rashid and Katja Hofmann and Jiang Bian},
      year={2025},
      eprint={2503.14070},
      archivePrefix={arXiv},
      primaryClass={cs.CV},
      url={https://arxiv.org/abs/2503.14070}, 
}

@misc{packcache,
      title={PackCache: A Training-Free Acceleration Method for Unified Autoregressive Video Generation via Compact KV-Cache}, 
      author={Kunyang Li and Mubarak Shah and Yuzhang Shang},
      year={2026},
      eprint={2601.04359},
      archivePrefix={arXiv},
      primaryClass={cs.CV},
      url={https://arxiv.org/abs/2601.04359}, 
}

@inproceedings{sito,
  title={Training-free and hardware-friendly acceleration for diffusion models via similarity-based token pruning},
  author={Zhang, Evelyn and Tang, Jiayi and Ning, Xuefei and Zhang, Linfeng},
  booktitle={Proceedings of the AAAI Conference on Artificial Intelligence},
  volume={39},
  number={9},
  pages={9878--9886},
  year={2025}
}

@misc{dato,
      title={Token Pruning for Caching Better: 9 Times Acceleration on Stable Diffusion for Free}, 
      author={Evelyn Zhang and Bang Xiao and Jiayi Tang and Qianli Ma and Chang Zou and Xuefei Ning and Xuming Hu and Linfeng Zhang},
      year={2024},
      eprint={2501.00375},
      archivePrefix={arXiv},
      primaryClass={cs.CV},
      url={https://arxiv.org/abs/2501.00375}, 
}

@misc{vbench,
      title={VBench: Comprehensive Benchmark Suite for Video Generative Models}, 
      author={Ziqi Huang and Yinan He and Jiashuo Yu and Fan Zhang and Chenyang Si and Yuming Jiang and Yuanhan Zhang and Tianxing Wu and Qingyang Jin and Nattapol Chanpaisit and Yaohui Wang and Xinyuan Chen and Limin Wang and Dahua Lin and Yu Qiao and Ziwei Liu},
      year={2023},
      eprint={2311.17982},
      archivePrefix={arXiv},
      primaryClass={cs.CV},
      url={https://arxiv.org/abs/2311.17982}, 
}

@misc{lpips,
      title={The Unreasonable Effectiveness of Deep Features as a Perceptual Metric}, 
      author={Richard Zhang and Phillip Isola and Alexei A. Efros and Eli Shechtman and Oliver Wang},
      year={2018},
      eprint={1801.03924},
      archivePrefix={arXiv},
      primaryClass={cs.CV},
      url={https://arxiv.org/abs/1801.03924}, 
}

@misc{3dvae,
      title={Wan: Open and Advanced Large-Scale Video Generative Models}, 
      author={Team Wan and Ang Wang and Baole Ai and Bin Wen and Chaojie Mao and Chen-Wei Xie and Di Chen and Feiwu Yu and Haiming Zhao and Jianxiao Yang and Jianyuan Zeng and Jiayu Wang and Jingfeng Zhang and Jingren Zhou and Jinkai Wang and Jixuan Chen and Kai Zhu and Kang Zhao and Keyu Yan and Lianghua Huang and Mengyang Feng and Ningyi Zhang and Pandeng Li and Pingyu Wu and Ruihang Chu and Ruili Feng and Shiwei Zhang and Siyang Sun and Tao Fang and Tianxing Wang and Tianyi Gui and Tingyu Weng and Tong Shen and Wei Lin and Wei Wang and Wei Wang and Wenmeng Zhou and Wente Wang and Wenting Shen and Wenyuan Yu and Xianzhong Shi and Xiaoming Huang and Xin Xu and Yan Kou and Yangyu Lv and Yifei Li and Yijing Liu and Yiming Wang and Yingya Zhang and Yitong Huang and Yong Li and You Wu and Yu Liu and Yulin Pan and Yun Zheng and Yuntao Hong and Yupeng Shi and Yutong Feng and Zeyinzi Jiang and Zhen Han and Zhi-Fan Wu and Ziyu Liu},
      year={2025},
      eprint={2503.20314},
      archivePrefix={arXiv},
      primaryClass={cs.CV},
      url={https://arxiv.org/abs/2503.20314}, 
}

@misc{vq1,
      title={Language Model Beats Diffusion -- Tokenizer is Key to Visual Generation}, 
      author={Lijun Yu and José Lezama and Nitesh B. Gundavarapu and Luca Versari and Kihyuk Sohn and David Minnen and Yong Cheng and Vighnesh Birodkar and Agrim Gupta and Xiuye Gu and Alexander G. Hauptmann and Boqing Gong and Ming-Hsuan Yang and Irfan Essa and David A. Ross and Lu Jiang},
      year={2024},
      eprint={2310.05737},
      archivePrefix={arXiv},
      primaryClass={cs.CV},
      url={https://arxiv.org/abs/2310.05737}, 
}

@misc{vq2,
      title={Image and Video Tokenization with Binary Spherical Quantization}, 
      author={Yue Zhao and Yuanjun Xiong and Philipp Krähenbühl},
      year={2024},
      eprint={2406.07548},
      archivePrefix={arXiv},
      primaryClass={cs.CV},
      url={https://arxiv.org/abs/2406.07548}, 
}

\newpage
\appendix
\setcounter{figure}{0}
\setcounter{table}{0}
\setcounter{page}{1}
\setcounter{linenumber}{1}

\renewcommand{\thefigure}{A.\arabic{figure}}
\renewcommand{\thetable}{A.\arabic{table}}
\renewcommand{\thepage}{A.\arabic{page}}

\begin{center}
{\Large \textbf{Appendix}}
\end{center}

\section{Attention Map and Important Tokens.}
\vspace{-0.2in}
\begin{figure}[H]
  \centering
  \includegraphics[width=\textwidth]{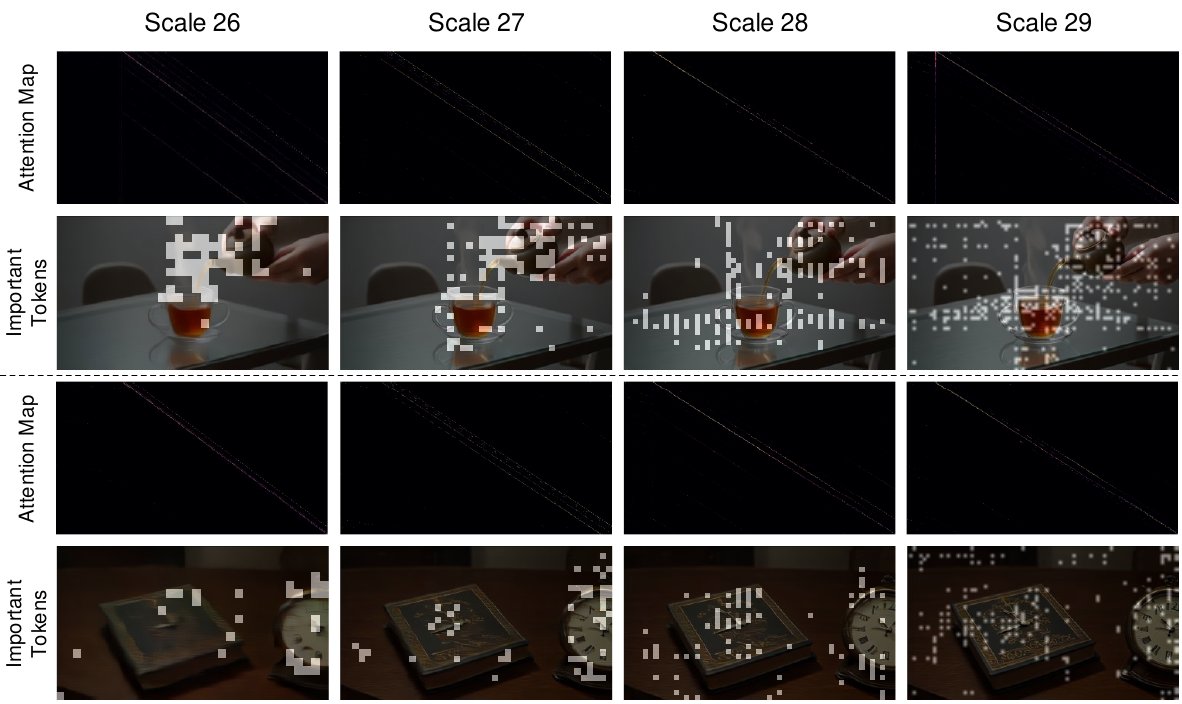}
  \caption{Visualization of attention score maps and their corresponding important tokens.}
  \label{suppl_attn_map}
\end{figure}

\noindent To further justify our pruning strategy, we analyze the relationship between the self-attention mechanism and the regions requiring refinement within the STAR transformer backbone. Specifically, we compute attention score maps by summing the attention scores over the query dimension.

\medskip

\noindent As illustrated in Fig.~\ref{suppl_attn_map}, these heatmaps exhibit a high degree of sparsity. Notably, the regions with high attention scores strongly align with the non-converged tokens—the areas that still require spatiotemporal updates and textural refinement. This alignment indicates that the self-attention mechanism naturally focuses on parts of the video that are not yet fully converged. By pruning tokens with near-zero attention weights, \methodname effectively concentrates computational power on these critical, non-converged regions. This optimization ensures that our spatiotemporal pruning remains consistent with the model's internal behavior, maintaining high-fidelity synthesis while eliminating redundant computations.

\newpage

\section{Profiling the Token Reduction Overhead.} \label{app:overhead}

\vspace{-0.2in}
\begin{figure}[H]
  \centering
  \includegraphics[width=\textwidth]{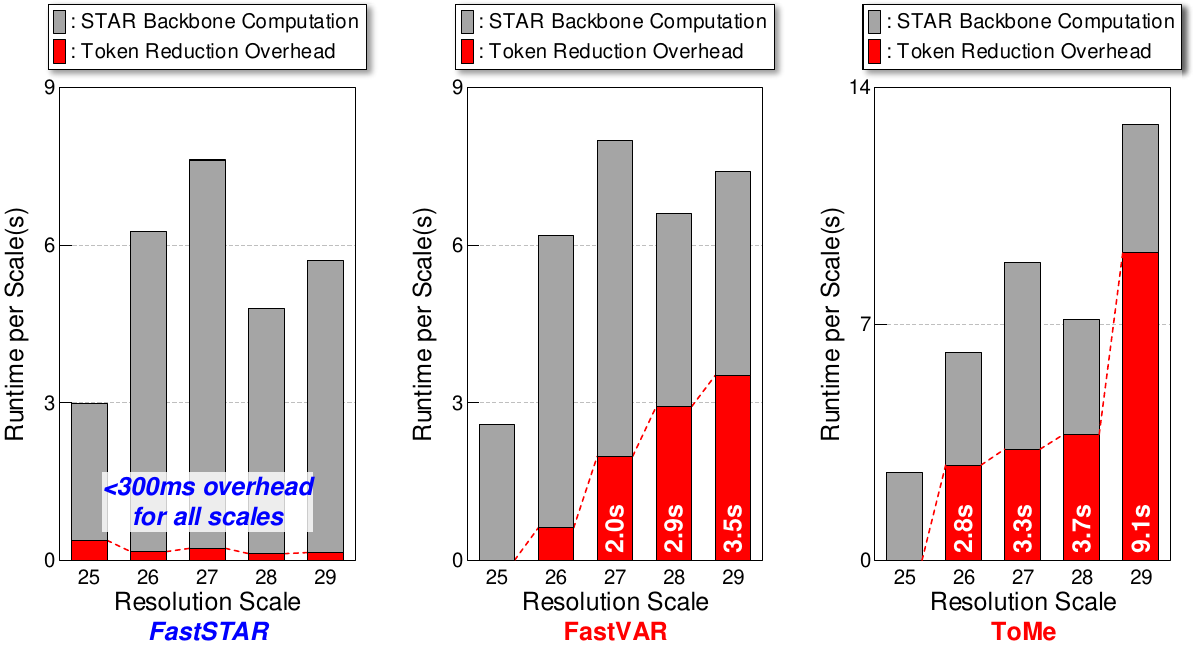}
  \caption{Latency breakdown and overhead analysis across different resolution scales.}
  \vspace{-0.2in}
  \label{suppl_overhead}
\end{figure}

\noindent To empirically validate the minimal overhead of our proposed STTP and PU mechanisms as discussed in \S~\ref{method:sttp}, we conducted precise latency profiling using NVIDIA Nsight Systems. As illustrated in Fig.~\ref{suppl_overhead}, we decomposed the runtime per resolution scale into two components: the STAR Backbone Computation (representing the original InfinityStar execution time) and the Token Reduction Overhead (representing the additional time required by each acceleration scheme). All measurements were recorded after a 30\% warmup phase to ensure architectural stability.

\medskip

\noindent The profiling results reveal a critical scalability bottleneck in prior methods. For FastVAR and ToMe, the overhead increases quadratically with the resolution scale, primarily due to layer-wise merging and unmerging operations within the transformer blocks. At the final scale (Scale 29), where the token count reaches 72,000, the overhead for ToMe exceeds 9.1s, significantly diminishing the practical gains from token reduction. In contrast, \methodname maintains a near-constant overhead of less than 300ms across all scales, regardless of the token counts. This efficiency stems from our similarity-based pruning strategy, which operates on the fixed, lower-dimensional video feature map rather than the transformer backbone's expanding token space. \methodname ensures that the reduction in backbone computation directly translates into end-to-end acceleration without introducing a new latency bottleneck.

\newpage

\section{Inference Efficiency on Another GPU.}
\begin{table}[!h]
  \centering
  \vspace{-0.3in}
  \includegraphics[width=\textwidth]{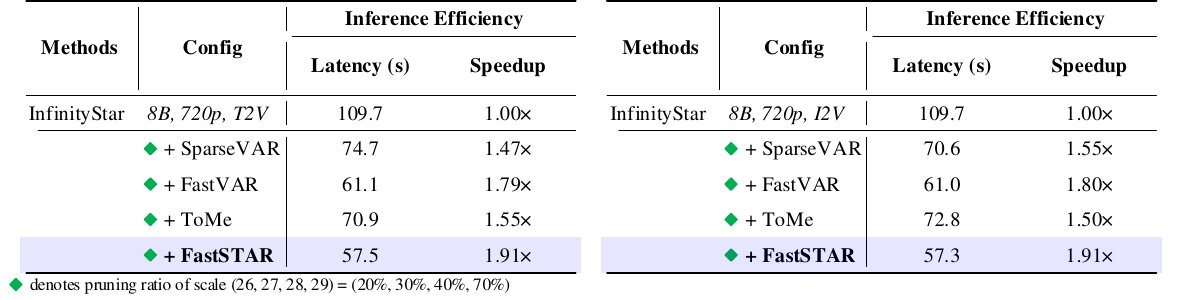}
  \caption{Inference efficiency benchmarking on A100 GPU (720p, 5s, 81 frames)}
  \label{suppl_a100}
  \vspace{-0.4in}
\end{table}

\noindent To demonstrate the hardware-agnostic efficiency of our framework, we evaluate the inference performance of \methodname on a single NVIDIA A100 GPU (80GB). As summarized in Table.~\ref{suppl_a100}, our method consistently achieves the highest speedup across both T2V and I2V tasks compared to all other baselines.

\medskip

\noindent In the T2V task, \methodname delivers a substantial speedup that outperforms established baselines, including SparseVAR, FastVAR, and ToMe. Similarly, in the I2V task, our method maintains a clear lead over existing reduction schemes, establishing the Pareto frontier. These results confirm that the spatiotemporal similarity-based pruning strategy is not limited to specific high-end hardware but delivers robust, consistent acceleration across different GPU architectures.

\section{Geometric and Statistical Analysis of the Score Fusion.}
\begin{figure}[!h]
  \centering
  \vspace{-0.3in}
  \includegraphics[width=\textwidth]{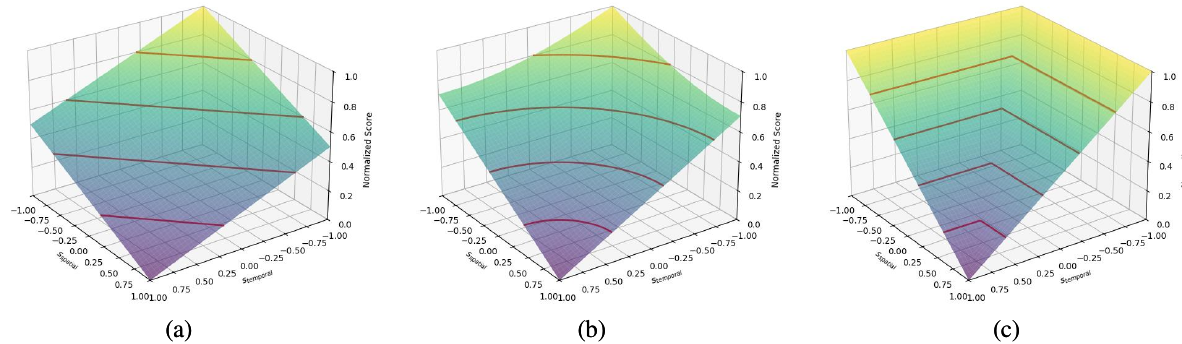}
  \caption{3D landscapes of the spatiotemporal score fusion function, illustrating the fusion geometries for (a) $p=1$ (Addition), (b) $p=2$ (Euclidean), and (c) $p=\infty$ (Maximum).}
  \label{suppl_norm}
  \vspace{-0.3in}
\end{figure}

\smallskip

\noindent To provide deeper intuition for our ablation study (\S~\ref{exp:ablation}) on the norm degree $p \in \{1, 2, \infty\}$, Fig.~\ref{suppl_norm} visualizes the 3D landscapes of our spatial and temporal similarity fusion functions.

\begin{itemize}
    \item \textbf{(a) $l_1$-norm ($p=1$):} Forms a linear, planar surface. It treats both metrics independently with uniform weighting, failing to emphasize tokens that are moderately important in both domains.
    \item \textbf{(b) $l_2$-norm ($p=2$):} Forms a smooth, convex landscape. This synergistic geometry effectively couples the two dimensions, acting as a \textit{balanced integrator} that smoothly amplifies tokens that exhibit variation across both domains. This corroborates our optimal empirical results of the highest SSIM and the lowest LPIPS.
    \item \textbf{(c) Max norm ($p=\infty$):} Defined by sharp creases, it bounds the score using only the maximum dimension. This ignores the secondary metric entirely, leading to suboptimal feature representation.
\end{itemize}

\noindent These geometries visually demonstrate why our $l_2$-norm-based method is the most robust choice for integrating spatiotemporal dynamics while preserving critical high-frequency details.

\begin{figure}[!h]
  \centering
  \vspace{-0.3in}
  \includegraphics[width=\textwidth]{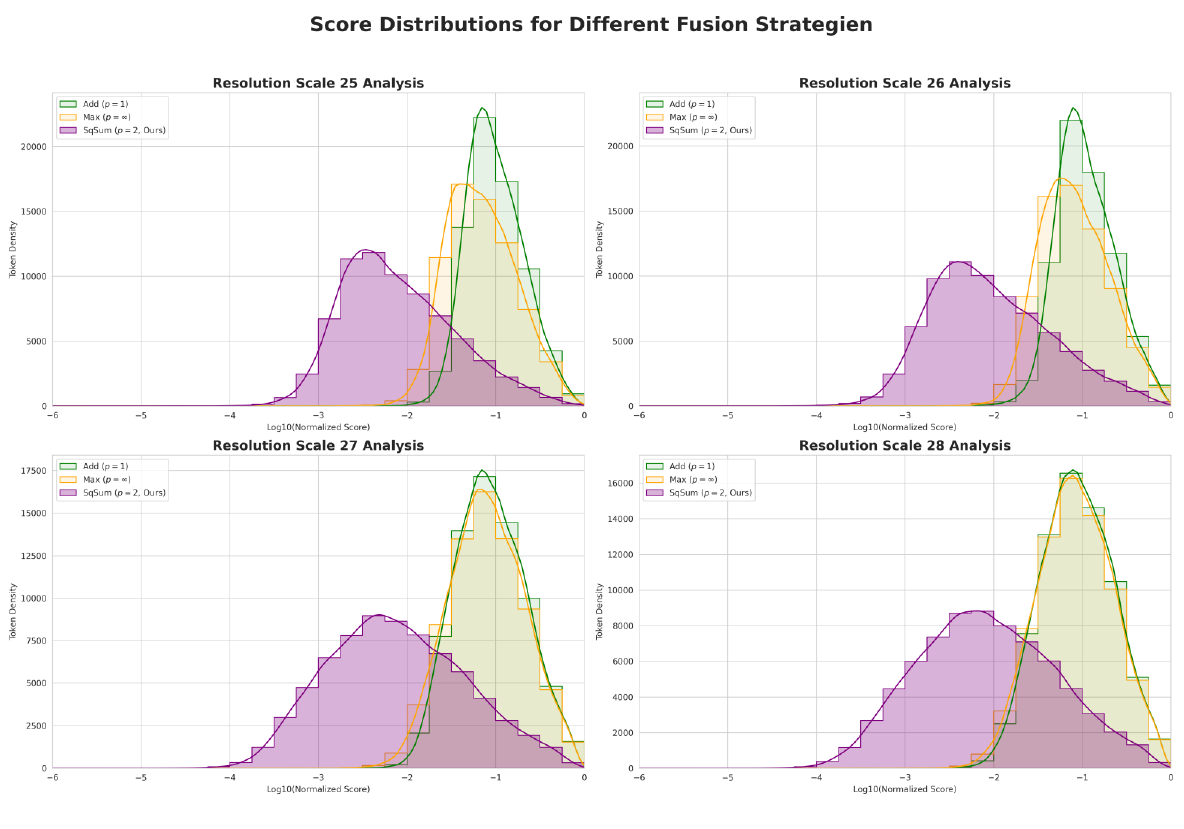}
  \caption{Log-scale spatiotemporal score distribution analysis of fusion functions across resolution scales.}
  \label{suppl_dist}
  \vspace{-0.4in}
\end{figure}

\medskip
\noindent \textbf{Score Distribution Analysis.} To further illustrate this advantage, Fig.~\ref{suppl_dist} visualizes the resulting token score distributions across varying resolution scales. The scores generated by $l_1$-norm (green) and Max norm (yellow) are highly concentrated within a narrow band. This dense clustering severely limits discriminative power, making it difficult to set clear boundaries during our top-$k$ masking process, as many tokens share near-identical scores. In contrast, our $l_2$-norm formulation (purple) produces a significantly broader distribution with a wider dynamic range. This expanded variance smoothly disperses the scores, enabling the top-$k$ selection mechanism to distinctly and robustly separate critical tokens from redundant ones.

\newpage

\section{More Quantitative Results.} \label{app:vbench_details}
\begin{table}[!h]
  \centering
  \vspace{-0.3in}
  \includegraphics[width=\textwidth]{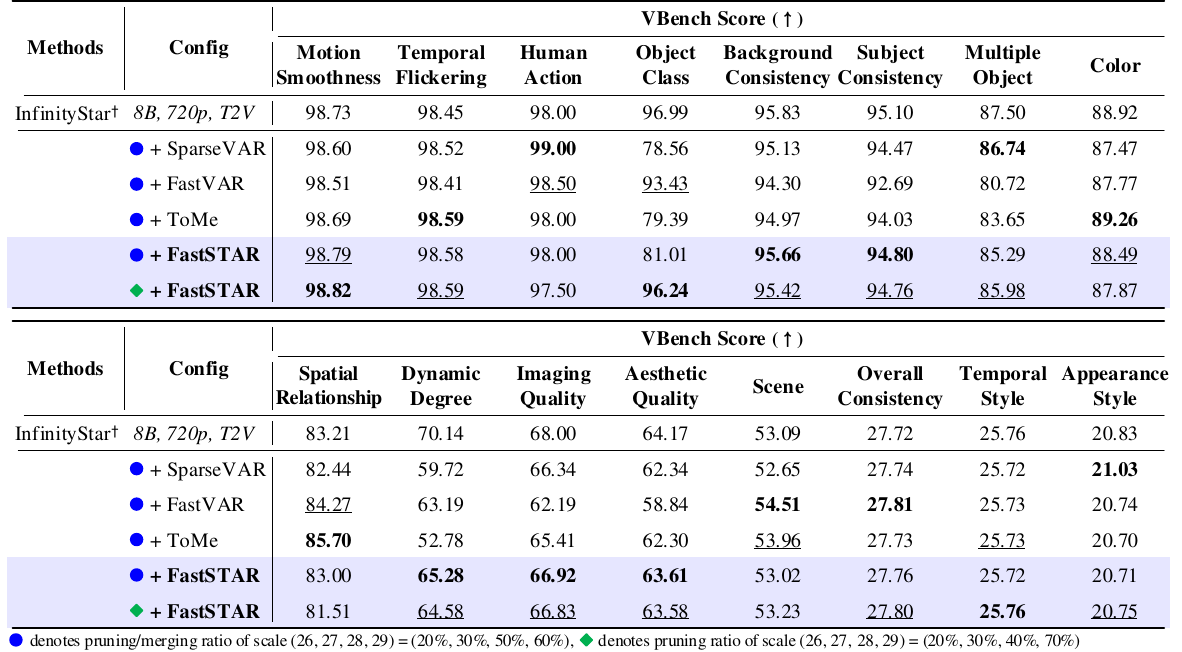}
  \caption{Quantitative and efficiency benchmarking for 720p, 5s (81 frames) Text-to-Video generation. \textbf{Best results} are highlighted in bold and \underline{second-best results} are underlined; $\dagger$ denotes results using prompt rewriting.}
  \vspace{-0.3in}
  \label{suppl_vbench_full}
\end{table}

\noindent As discussed in \S~\ref{exp:vbench}, we provide a comprehensive breakdown of the VBench scores to offer a more granular evaluation of our framework. Table~\ref{suppl_vbench_full} presents the results across all 16 dimensions of the VBench suite for the 720p, T2V generation task. These scores were obtained under the same experimental settings and evaluation protocols as those detailed in the main manuscript to ensure consistency across all performance metrics. The full report further substantiates the robustness of \methodname in maintaining high fidelity across diverse categories, ranging from motion smoothness to appearance style.

\vspace{-0.1in}

\section{More Qualitative Results.}
\begin{figure}[H]
  \centering
  \includegraphics[width=\textwidth]{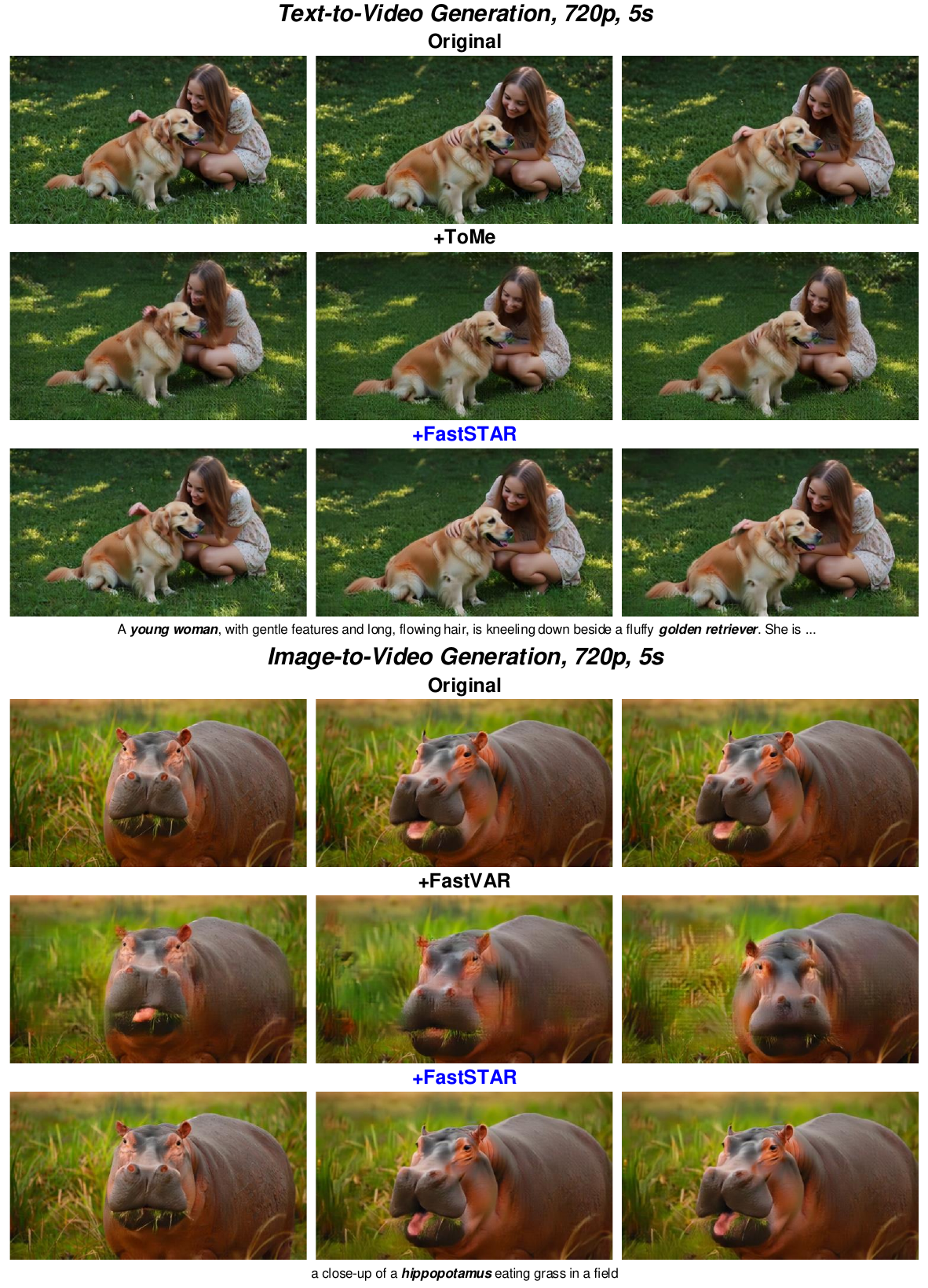}
  \caption{Additional Side-by-Side comparison of FastSTAR with ToMe, FastVAR.}
  \vspace{-0.2in}
  \label{suppl_quality_compare}
\end{figure}

\begin{figure}[H]
  \centering
  \includegraphics[width=\textwidth]{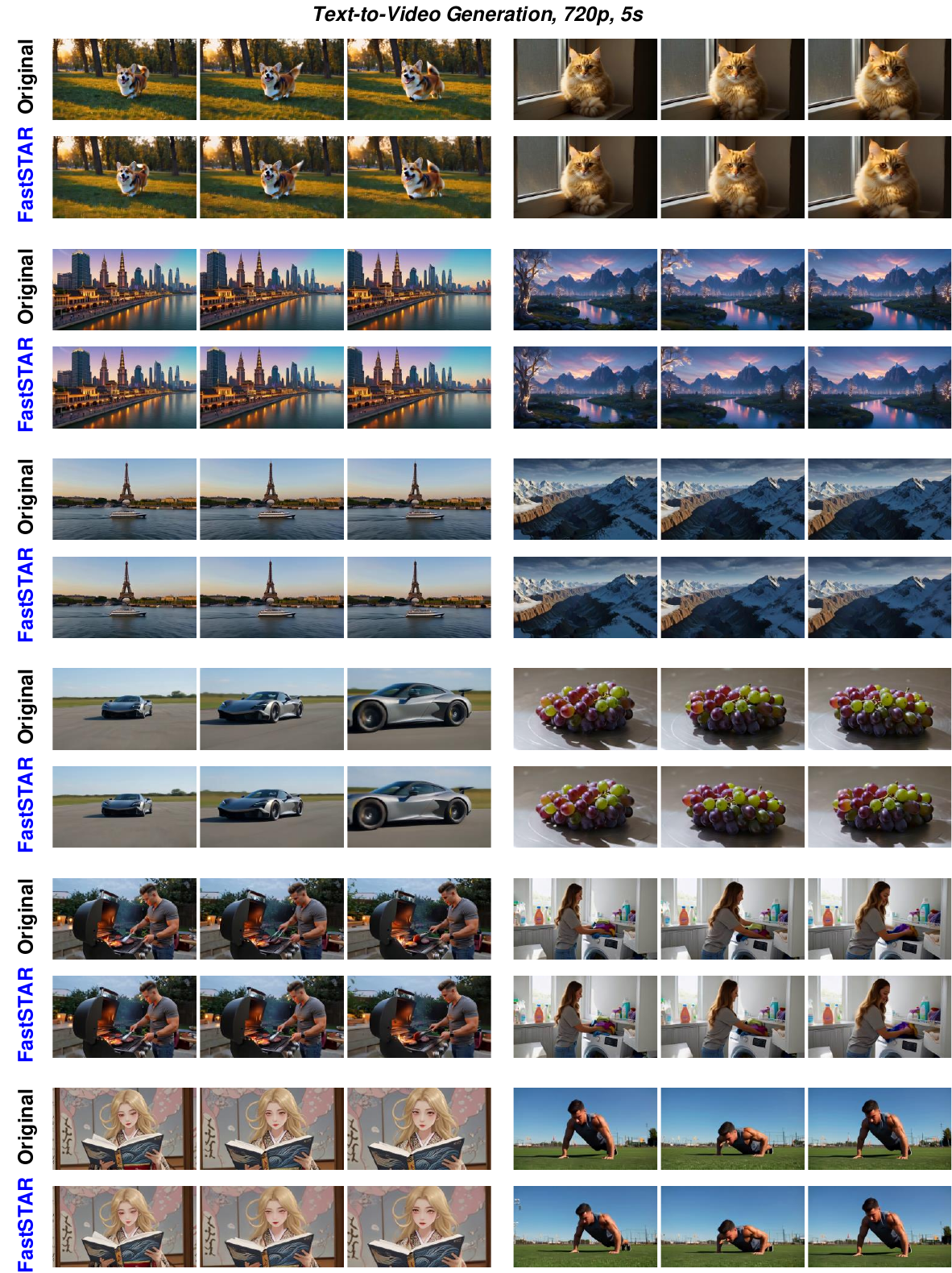}
  \caption{Text-to-Video generation results at 720p resolution.}
  \vspace{-0.2in}
  \label{suppl_t2v_720}
\end{figure}

\begin{figure}[H]
  \centering
  \includegraphics[width=\textwidth]{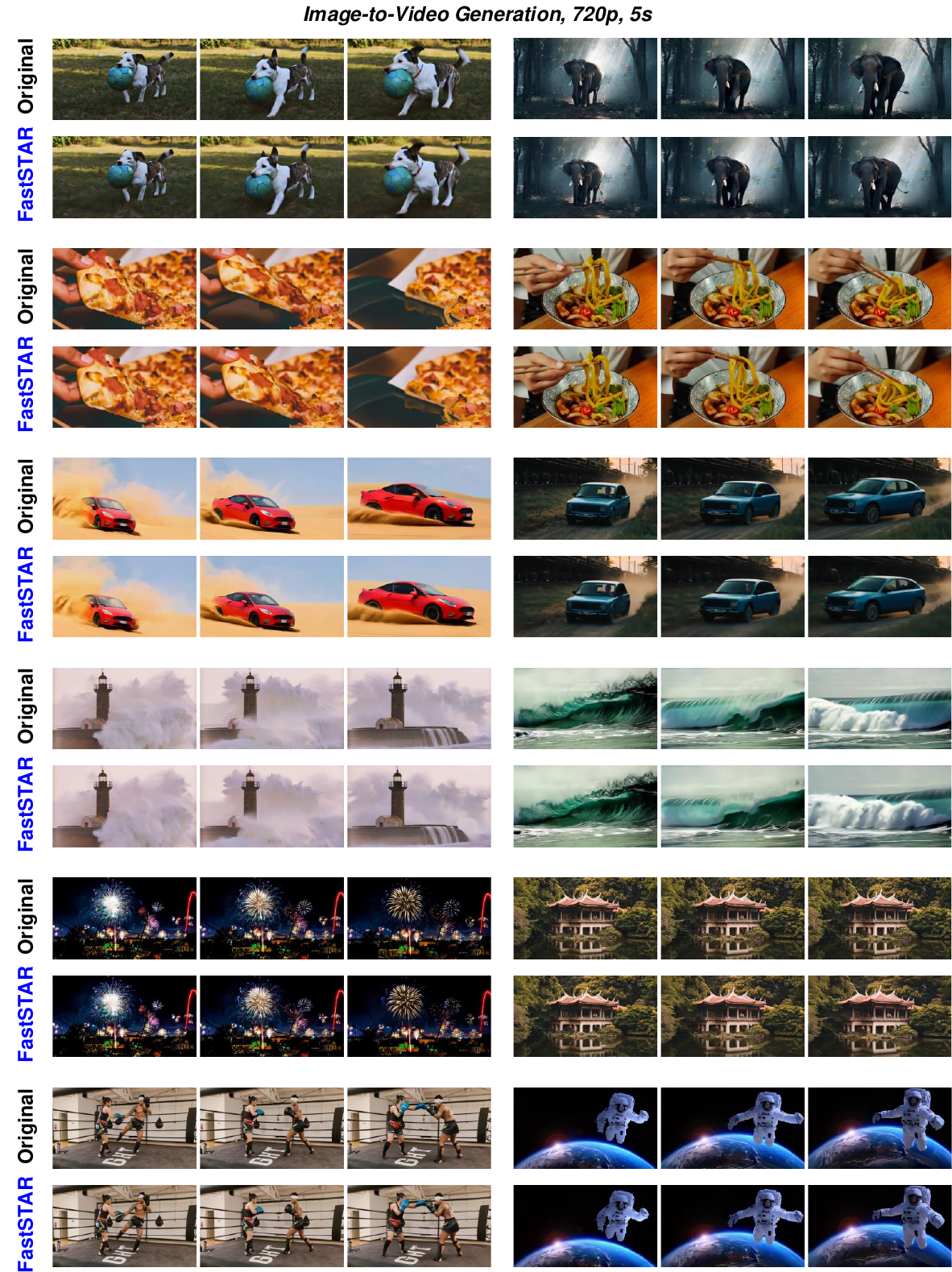}
  \caption{Image-to-Video generation results at 720p resolution.}
  \vspace{-0.2in}
  \label{suppl_i2v_720}
\end{figure}

\noindent To provide a more comprehensive evaluation of \methodname, we present extended qualitative results across diverse scenarios and configurations. Fig.~\ref{suppl_quality_compare} offers additional side-by-side comparisons, highlighting our method's superior ability to preserve fine-grained textures compared to existing baselines. The galleries in Fig.~\ref{suppl_t2v_720} and Fig.~\ref{suppl_i2v_720} demonstrate the robust generation capabilities of our framework in 720p Text-to-Video and Image-to-Video tasks, showcasing consistent high-fidelity synthesis across various semantic categories.

\begin{figure}[H]
  \centering
  \includegraphics[width=\textwidth]{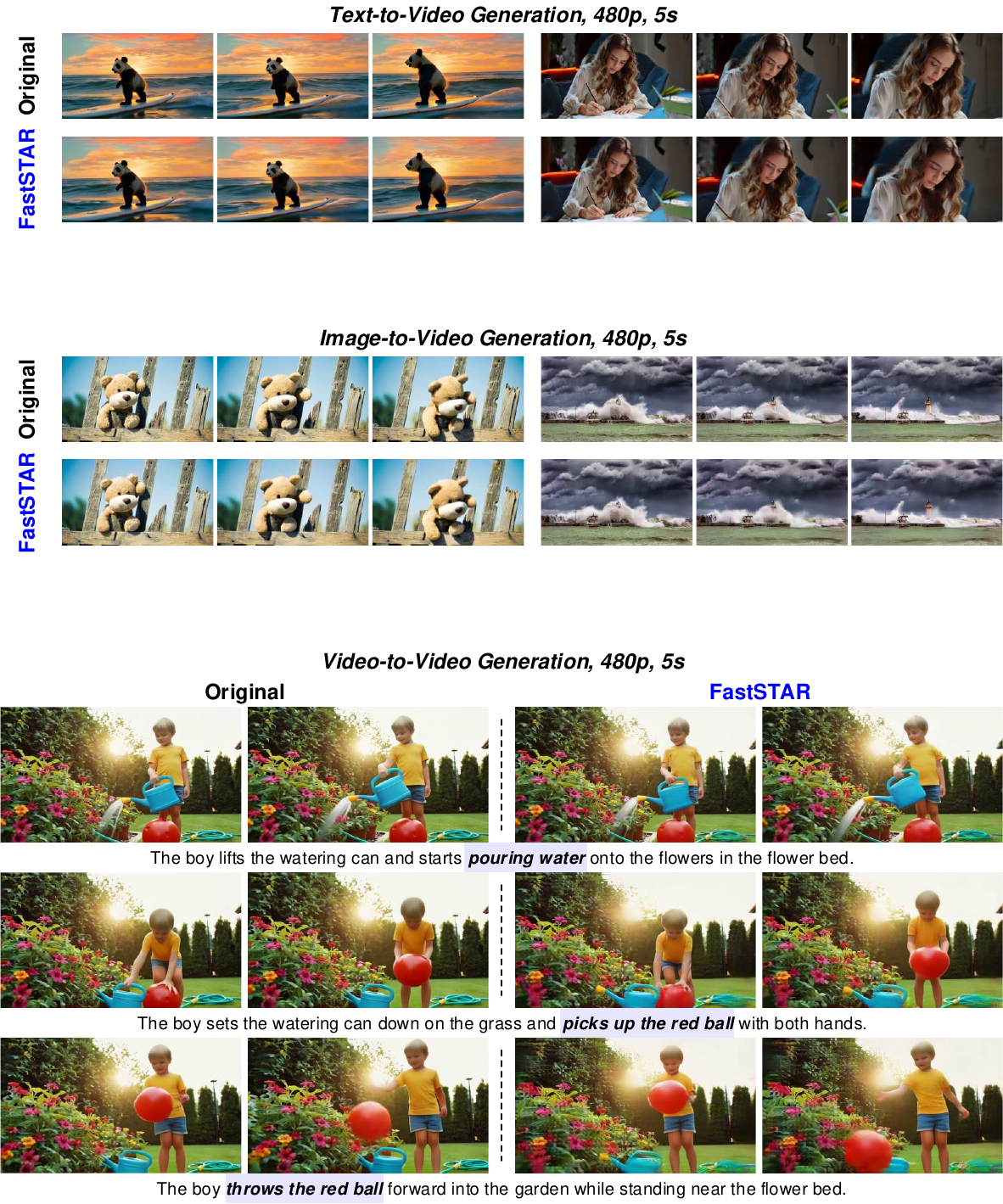}
  \caption{Qualitative examples across multiple 480p generation tasks.}
  \vspace{-0.2in}
  \label{suppl_480}
\end{figure}

\noindent Furthermore, Fig.~\ref{suppl_480} illustrates the versatility of \methodname in 480p resolution tasks, including T2V, I2V, and V2V. These results collectively confirm that our spatiotemporal pruning strategy effectively maintains structural integrity and motion continuity across different resolutions and conditioning signals, producing high-fidelity results from the original while significantly reducing inference latency.

\newpage

\vspace{-0.2in}

\newpage
\vspace{-0.2in}

\newpage
\vspace{-0.2in}

\newpage
\vspace{-0.2in}
 \label{app:480p_qualitative}

\end{document}